\documentclass[10pt,twocolumn]{article}
\usepackage{float}

\usepackage[margin=0.72in,columnsep=0.25in]{geometry}
\usepackage{times}
\usepackage{microtype}
\usepackage{graphicx}
\usepackage{booktabs}
\usepackage{multirow}
\usepackage{array}
\usepackage{amsmath,amssymb,bm}
\usepackage{mathtools}
\usepackage{algorithm}
\usepackage{algpseudocode}
\usepackage{enumitem}
\usepackage{xcolor}
\usepackage{url}
\usepackage{hyperref}
\usepackage{cleveref}
\usepackage{tabularx}
\usepackage{threeparttable}
\usepackage{siunitx}
\usepackage{tikz}
\usetikzlibrary{arrows.meta,positioning,fit,calc}
\usepackage{balance}
\usepackage{subcaption}
\usepackage{placeins}
\usepackage{longtable}
\newcommand{\cmark}{\checkmark}
\usepackage{xspace}
\usepackage[backend=biber,style=numeric,sorting=none,maxbibnames=6,giveninits=true]{biblatex}
\hypersetup{colorlinks=true,citecolor=blue,linkcolor=blue,urlcolor=blue}
\newcommand{\model}{\textbf{TAP-Path}\xspace}
\newcommand{\vir}{Virchow2\xspace}

\newcommand{\ece}{\mathrm{ECE}}

\title{\textbf{TAP-Path: Task-Adaptive Structural and Token Pruning for Efficient and Trustworthy Pathology Foundation Models}}
\author{
\textbf{Mehedi Hasan}$^{1}$ \quad
\textbf{Ashfak Yeafi}$^{2}$ \quad
\textbf{Md Khairul Islam}$^{3}$\\[2pt]
$^{1}$Department of Computer Science and Engineering, Brac University, Dhaka, Bangladesh\\
$^{2}$EEE, Khulna University of Engineering \& Technology, Khulna, Bangladesh\\
$^{3}$Mathematics and Computer Science, Hobart and William Smith Colleges, Geneva, NY, USA\\[2pt]
\footnotesize
\texttt{mehedi.hasan1@g.bracu.ac.bd} \quad
\texttt{yeafiashfak@gmail.com} \quad
\texttt{khairul.robotics@gmail.com}
}

\date{}
\begin{document}
\maketitle

\begin{abstract}
Pathology foundation models have substantially improved transferable representation learning for histopathology, but recent gains increasingly rely on encoders with hundreds of millions of parameters and high inference cost. This creates a practical mismatch between representational scale and task-specific deployment, especially when a downstream application requires only a subset of the pretrained hierarchy. We propose \model, a task-adaptive compression framework that directly restructures a pretrained \vir encoder rather than distilling it into a separate student. \model combines validation-driven transformer-block selection, physical removal of redundant blocks, input-adaptive patch-token pruning, multi-depth feature recovery, and a lightweight gated task head. The final model retains 24 of 32 transformer blocks and 70\% of patch tokens after the pruning point, reducing encoder parameters by 24.96\% (631.24M to 473.70M) and analytical encoder compute by 35.20\% (340.13G to 220.40G FLOPs). Across three task-head optimization seeds, \model achieved 87.98$\pm$0.067\% test accuracy, 81.26$\pm$0.49\% balanced accuracy, and 82.38$\pm$0.48\% macro-F1 on a 32-class histopathology benchmark, slightly exceeding full \vir (86.89\% accuracy) and UNI2-h (87.67\%) while using fewer deployed parameters and substantially less analytical compute. Beyond predictive performance, we evaluate calibration, probabilistic quality, failure detection, rare-class behavior, and selective prediction. \model obtained a Brier score of 0.1800$\pm$0.0005 and failure-detection AUROC of 0.9047$\pm$0.0060. A validation-only rare-aware objective increased rare-class balanced accuracy in a secondary operating analysis, providing a complementary operating point that characterizes the trade-off between overall accuracy and minority-class sensitivity. Frozen external evaluation on 433 CPTAC samples yielded 91.22$\pm$0.83\% accuracy and 91.10$\pm$0.81\% balanced accuracy. These results indicate that task-adaptive structural and token sparsification can move large pathology foundation models toward a more favorable accuracy--efficiency operating point while preserving measurable reliability characteristics under internal and external evaluation.
\end{abstract}

\noindent\textbf{Keywords:} Computational pathology; pathology foundation models; structural pruning; token pruning; trustworthy AI; model efficiency.

\section{Introduction}

Digital pathology has moved rapidly from task-specific convolutional pipelines toward large pretrained vision and vision--language foundation models. Models such as UNI, CONCH, Virchow, Prov-GigaPath, CHIEF, Hibou, GPFM, and PathOrchestra have demonstrated that large-scale self-supervised, weakly supervised, or multimodal pretraining can produce reusable histopathology representations across cancer classification, biomarker prediction, retrieval, prognosis, and other downstream tasks \cite{chen2024uni,lu2024conch,vorontsov2024virchow,xu2024gigapath,wang2024chief,nechaev2024hibou,ma2025gpfm,yan2025pathorchestra}. The emerging consensus, however, is not that a single foundation model is uniformly superior. Independent benchmarking has shown substantial task dependence, cohort dependence, sensitivity to adaptation strategy, and variation under external evaluation \cite{campanella2025benchmark,lee2025benchmark,neidlinger2025benchmark,bareja2026benchmark}. This makes deployment efficiency and reliability increasingly important alongside raw predictive performance.

The computational cost of contemporary pathology foundation models is particularly relevant. From a systems perspective, this mismatch is amplified at whole-slide scale. A tissue-rich WSI may generate thousands to tens of thousands of candidate tiles at diagnostic magnification, and the encoder is invoked repeatedly before slide-level aggregation or downstream decision logic. A reduction of one third in per-patch transformer compute therefore compounds across the entire WSI rather than saving only a single forward pass. This observation motivates task-specific restructuring of the patch encoder itself, rather than limiting efficiency analysis to the number of trainable downstream parameters. \vir uses a ViT-H/14-scale encoder with approximately 632M parameters \cite{vorontsov2024virchow}, while other leading models are similarly large. Whole-slide applications amplify this burden because a single slide may generate hundreds or thousands of patches. Even when a large encoder is frozen, its stored parameters and forward-pass operations remain present at inference. Parameter-efficient fine-tuning therefore reduces optimization cost but does not necessarily reduce the resident encoder or patch-level inference burden. Recent medical-image studies have increasingly recognized lightweight design as a distinct problem from parameter-efficient adaptation \cite{kuang2025lwctrans,silva2025peft,islam2025involution}. In computational pathology, SUDA explicitly targets this problem by distilling heavy foundation models into lightweight students \cite{zhong2026suda}. Distillation can be highly effective, but it changes the model family, requires a teacher--student transfer procedure, and may need to be repeated for new tasks or domains. This leaves an important complementary question: \emph{how much of an already pretrained large pathology encoder is actually necessary for a specific downstream task?}

Generic vision research suggests that substantial redundancy can exist both along transformer depth and within the token sequence. Structured sparsification and token-pruning approaches reduce computation by removing blocks, tokens, or interactions that contribute little to a target task \cite{liu2024tokenpruning,xu2024gtpvit,bergner2025tokencropr,an2025sparsevit,marchetti2025token}. Nevertheless, pathology foundation models are not ordinary ImageNet classifiers. Histopathological discriminative evidence may occupy small tissue regions, vary across magnification and morphology, and be especially sparse for low-prevalence classes. Aggressive pruning that looks attractive from a FLOP perspective can therefore erase diagnostically useful evidence. The desired objective is not maximum compression alone, but a validated Pareto operating point that preserves task utility while reducing genuine encoder cost. The design philosophy is similar to recent efficiency-oriented vision architectures that explicitly optimize accuracy against parameters and compute \cite{munir2025adaptvig}, but here the problem is addressed by restructuring a pretrained pathology foundation model rather than designing a new backbone from scratch.

A second limitation of efficiency-only evaluation is that a compressed medical model can retain accuracy while becoming less trustworthy in ways that aggregate accuracy does not reveal. Calibration error, negative log-likelihood, Brier score, confidence-based failure detection, and selective risk quantify different aspects of whether predicted probabilities are useful for decision support \cite{guo2017calibration,ovadia2019uncertainty,geifman2019selective}. Recent pathology work has likewise emphasized domain robustness, fairness, and vulnerability to non-biological scanner or laboratory variation \cite{huang2025flex,komen2026robust,jahanifar2025domain}. Thus, a credible compression study in computational pathology should establish not only accuracy and compute reduction but also whether uncertainty remains informative, minority classes are characterized explicitly, and the locked model generalizes to an independent cohort.

This study addresses these gaps with \model, short for \emph{TAP-Path, a task-adaptive structural and token pruning framework for pathology foundation models}. \model begins from the pretrained \vir encoder and performs validation-only structural analysis to identify a task-relevant subset of transformer blocks. Unlike freezing or masking, selected blocks are physically retained in the deployed encoder while unselected blocks are removed. The structurally compressed model is then paired with input-adaptive token pruning that preserves the prefix token and retains the most task-relevant patch tokens using a combined class-token similarity and feature-magnitude score. To compensate for information loss from shortening both depth and token sequence, \model collects multi-depth features from four locations in the retained hierarchy and fuses them through a learned soft gate before classification. The model is selected entirely on training and validation data; the internal test set and CPTAC external cohort are evaluated only after configuration locking. The final single-head model is confirmed with three random seeds and evaluated with accuracy, balanced accuracy, macro-F1, rare-class balanced accuracy, calibration, probabilistic losses, failure-detection AUROC, runtime, parameters, analytical FLOPs, and frozen external performance.

In this paper, we propose \model, a task-adaptive compression framework that restructures a pretrained pathology foundation model for efficient and reliable downstream deployment. 

The main contributions of this study are summarized as follows:
\begin{itemize}[leftmargin=*,itemsep=3pt,topsep=3pt]

\item We propose \textbf{\model}, a task-adaptive compression framework that identifies task-relevant transformer depth and physically reconstructs a compact, non-contiguous subnetwork from the pretrained \vir hierarchy. This enables direct compression of the foundation model without training or distilling a separate student network.

\item We develop an \textbf{adaptive token-pruning and multi-depth recovery strategy} that reduces redundant patch-token computation while preserving complementary representations from multiple stages of the compressed hierarchy. A learned gating mechanism integrates these representations for downstream prediction.

\item We systematically investigate the \textbf{computational efficiency of pathology foundation models} through deployed parameter count and analytical FLOPs, together with predictive performance. The proposed design reduces \vir encoder parameters by 24.96\% and FLOPs by 35.20\%, demonstrating that substantial computational redundancy can be removed while preserving competitive performance on the 32-class pathology task.

\item We conduct a \textbf{reliability-oriented evaluation relevant to trustworthy medical AI} by examining calibration, failure awareness, selective prediction, rare-class behavior, and statistical uncertainty. We further evaluate the locked model on two independent CPTAC cohorts without external adaptation or threshold tuning to assess generalization beyond the development data.


\end{itemize}





Together, these results show that task-specific structural and token redundancy can be removed from a large pathology foundation model while retaining competitive discrimination and informative uncertainty.

\section{Related Work}

\subsection{Pathology foundation models}

The current generation of computational-pathology foundation models has been driven by scale, data diversity, and self-supervised learning. UNI established a general-purpose pathology representation benchmark using large-scale self-supervision \cite{chen2024uni}; CONCH extended the paradigm to paired pathology image--text learning \cite{lu2024conch}; Virchow showed the value of million-slide-scale pretraining and a ViT-H backbone for common and rare cancers \cite{vorontsov2024virchow}; Prov-GigaPath introduced a whole-slide hierarchy trained on 1.3 billion tiles \cite{xu2024gigapath}; CHIEF combined tile-level and slide-level pretraining across multiple institutions \cite{wang2024chief}; and Hibou provided open DINOv2-based pathology encoders trained on more than one million WSIs \cite{nechaev2024hibou}. Subsequent models have broadened clinical task coverage and pretraining strategies, including BEPH, GPFM, and PathOrchestra \cite{yang2025beph,ma2025gpfm,yan2025pathorchestra}.

As the number of foundation models has grown, independent evaluation has become as important as model creation. Campanella et al. benchmarked public self-supervised pathology foundation models on clinical tasks \cite{campanella2025benchmark}. Lee et al. systematically examined adaptation strategies and data-limited scenarios, finding that the preferred adaptation procedure depends on the downstream setting \cite{lee2025benchmark}. Neidlinger et al. evaluated 19 foundation models across external clinical cohorts and found that relative model ranking changes with the task and data regime \cite{neidlinger2025benchmark}. A broader 2026 benchmark further compared general and pathology-specific vision and vision--language foundation models across multiple cohorts and task categories \cite{bareja2026benchmark}. These studies motivate controlled same-task comparisons rather than relying on headline performance reported on unrelated benchmarks. Table~\ref{tab:litcompare} summarizes the most directly relevant foundation-model, adaptation, and pruning studies and highlights the methodological gap addressed by TAP-Path.

The literature also increasingly separates representation quality from domain robustness. Knowledge-guided adaptation has been proposed to improve cross-domain generalization and fairness \cite{huang2025flex}, while recent work demonstrates that pathology foundation models may encode scanner and laboratory signatures that can compromise robustness \cite{komen2026robust}. Domain-generalization surveys similarly emphasize explicit external evaluation and transparent reporting of data sources \cite{jahanifar2025domain}. Our work follows this direction by locking the compressed architecture before independent CPTAC evaluation and by reporting probability-quality and failure-detection measures in addition to class prediction.

\subsection{Efficient adaptation and compression}

Several strategies can make foundation models easier to use downstream. Prompt tuning, adapters, and related parameter-efficient methods update only a small portion of the model, reducing optimization memory and the number of trainable weights. For example, prompt-guided adaptive model transformation has been proposed for pathology classification using representative patch sampling, visual prompts, and adapters \cite{lin2026pamt}; broader medical-image work has also demonstrated few-shot parameter-efficient tuning \cite{silva2025peft}. These approaches are useful when fine-tuning cost is the main bottleneck, but a frozen or adapter-tuned 600M-parameter encoder still executes most of the original forward pass.

Knowledge distillation instead transfers knowledge to a smaller student. GPFM uses a unified expert/self-distillation framework during pretraining \cite{ma2025gpfm}, and SUDA directly targets efficient pathology analysis by combining unsupervised distillation and domain adaptation; its student can approach the teacher with a small fraction of the parameters \cite{zhong2026suda}. Distillation is therefore a strong point of comparison, but it answers a different question from ours. \model asks whether a large pretrained encoder can be \emph{restructured in place} for a target task without building a separate student, thereby preserving the original representation family while physically eliminating task-redundant computation.

This distinction matters because resource efficiency has become a first-class objective in medical imaging. Lightweight hybrid architectures have reduced model size in segmentation and histopathology-specific applications \cite{kuang2025lwctrans,islam2025involution}. One study by \cite{islam2025involution}, explicitly evaluated model size, FLOPs, and MACs alongside predictive performance, illustrating the importance of quantitative efficiency assessment in histopathology models. Similarly, recent efficiency-oriented vision work such as AdaptViG evaluates a Pareto frontier between accuracy, parameters, and compute \cite{munir2025adaptvig}. We adopt the same scientific principle---efficiency must be demonstrated quantitatively---but apply it to the task-specific compression of a pathology foundation model.

\subsection{Transformer sparsification}

The transformer architecture offers at least two complementary axes of redundancy: depth and tokens. Structured methods remove layers, heads, or channels, while token-pruning methods reduce the sequence length dynamically. Vision-specific token pruning has been revisited for dense prediction \cite{liu2024tokenpruning}, graph-based propagation has been proposed to preserve token information under reduction \cite{xu2024gtpvit}, and Token Cropr demonstrated task-aware token removal across several vision problems \cite{bergner2025tokencropr}. Sparse structure exploration and learned token-scoring networks further show that efficiency can be improved by searching for task-relevant computation rather than uniformly shrinking the network \cite{an2025sparsevit,marchetti2025token}.

Recent pathology-specific pruning evidence further supports the premise that transformer redundancy can be removed selectively. Boudissa et al. analyzed attention-head similarity and confidence in histopathology ViTs and showed that pruning redundant heads can preserve or even improve classification performance while reducing computational burden \cite{boudissa2025headpruning}. This work is complementary to TAP-Path: their unit of sparsification is the attention head in a distilled ViT, whereas our method physically selects non-contiguous transformer blocks and subsequently prunes patch tokens inside a large pretrained pathology foundation model. The distinction strengthens the motivation for testing structural redundancy directly in pathology-specific transformers.

Pathology introduces a special constraint: the discriminative region may be spatially sparse. Token removal therefore has to be conditioned on feature content, and the pruning ratio should be chosen on a locked validation protocol rather than assumed a priori. \model combines a block-level task relevance signal with an input-adaptive token score, then uses multi-depth feature recovery to preserve information from distinct stages of the compressed hierarchy.

\subsection{Reliability and selective prediction}

Modern neural networks can be accurate yet poorly calibrated, motivating temperature scaling and expected calibration error (ECE) \cite{guo2017calibration}. Under dataset shift, uncertainty estimates can degrade even when in-distribution metrics appear strong \cite{ovadia2019uncertainty}. Selective prediction addresses a complementary practical question: if the system abstains on its least-confident cases, how quickly does error risk decrease \cite{geifman2019selective}? These concepts are particularly relevant to computational pathology, where externally acquired slides may differ in staining, preparation, scanner hardware, and population characteristics. Our operational trustworthiness analysis covers calibration (ECE), proper scoring rules (NLL and Brier), failure-detection AUROC, rare-class behavior, and risk--coverage. Together, these measures provide complementary evidence of probability quality, error awareness, selective-prediction behavior, and reliability under external evaluation. Their clinical interpretation requires prospective validation under the intended workflow.

\begin{table*}[t]
\centering
\caption{Representative recent pathology foundation-model studies most relevant to TAP-Path.}
\label{tab:litcompare}
\small
\resizebox{\textwidth}{!}{%
\begin{tabular}{lllll}
\toprule
Study & Year & Main strategy & Efficiency / adaptation & Gap addressed by TAP-Path\\
\midrule
UNI \cite{chen2024uni} & 2024 & General pathology FM & Frozen transfer & No task-specific structural compression\\
CONCH \cite{lu2024conch} & 2024 & Vision--language FM & Compact transferable encoder & No in-place block/token pruning\\
Virchow \cite{vorontsov2024virchow} & 2024 & Large-scale pathology FM & Scale-driven pretraining & High inference footprint\\
Prov-GigaPath \cite{xu2024gigapath} & 2024 & WSI foundation modeling & Hierarchical slide modeling & Focuses context, not encoder compression\\
Lee et al. \cite{lee2025benchmark} & 2025 & PFM adaptation benchmark & PEFT / few-shot adaptation & Trainable-efficiency, not physical reduction\\
SUDA \cite{zhong2026suda} & 2026 & Efficient pathology adaptation & Teacher--student distillation & Requires a separate student\\
Boudissa et al. \cite{boudissa2025headpruning} & 2025 & Histology ViT pruning & Attention-head pruning & Head-level, not PFM block+token pruning\\
PAMT \cite{lin2026pamt} & 2026 & Prompt-guided adaptation & Prompts / adapters & Full backbone largely retained\\
\textbf{TAP-Path} & -- & \textbf{Task-adaptive PFM compression} & \textbf{Physical block + adaptive token pruning} & \textbf{Efficiency + reliability + frozen external validation}\\
\bottomrule
\end{tabular}%
}
\end{table*}

\section{Materials and Methods}

\subsection{Study design}

The study followed a sequential model-selection and locked-evaluation protocol. Architecture screening and token-retention selection were performed exclusively using the training and validation partitions. The internal test set was not used for transformer-block selection, token-retention selection, loss-function selection, or calibration-hyperparameter optimization. After fixing the structural and token configuration, three independent task heads were trained using seeds 42, 123, and 2026. All external evaluations were subsequently performed using the locked configuration without external model adaptation or threshold optimization. This protocol maintains a strict separation between model development, internal testing, and external evaluation.

The internal benchmark comprised 25,495 image-level records representing 32 cancer classes derived from The Cancer Genome Atlas (TCGA) \cite{tcga2013pancancer}. The dataset was partitioned into 17,769 training, 3,867 validation, and 3,859 test images, with each image represented using the final 12-patch protocol. Table~\ref{tab:data} summarizes the resulting data partitions.

For independent external evaluation, we used 433 whole-slide images (WSIs) from the Clinical Proteomic Tumor Analysis Consortium (CPTAC), comprising 209 clear cell renal cell carcinoma (CCRCC) images \cite{clark2019ccrcc} and 224 uterine corpus endometrial carcinoma (UCEC) images \cite{dou2020ucec}. External evaluation was conducted at the image level using the locked model without external fitting or threshold optimization. Because the external cohorts represent only two of the 32 cancer classes included in the internal benchmark, external balanced accuracy was computed over the classes represented in CPTAC; metrics requiring averaging across all 32 internal classes were not used as primary measures of class-balanced external performance.

\begin{table}[t]
\centering
\caption{Dataset partitions used in the locked evaluation protocol.}
\label{tab:data}
\small
\begin{tabular}{lrr}
\toprule
Partition & Images & Classes \\
\midrule
Training & 17,769 & 32\\
Validation & 3,867 & 32\\
Internal test & 3,859 & 32\\
\midrule
Internal total & 25,495 & 32\\
External CPTAC & 433 & 2 present classes\\
\bottomrule
\end{tabular}
\end{table}

\subsection{Preprocessing and patch selection}
\label{sec:patchprotocol}

The internal representation is image-level but is deliberately constructed from multiple pathology regions. The final protocol uses a fixed 12-patch bag for every record. Patch selection was completed before TAP-Path architecture search and therefore does not use test-set model outcomes. Candidate regions were generated at multiple source crop sizes (224, 448, and 896 pixels) and subsequently resized to the encoder input resolution. To preserve scale coverage, the selector first reserves one high-confidence candidate from every available scale. Two additional high-entropy candidates are then forced into the bag so that difficult or ambiguous morphology is not systematically discarded. Remaining positions are filled by a combined utility criterion
\begin{equation}
q_i = 0.50\,c_i + 0.30\,d_i + 0.20\,h_i,
\label{eq:patchutility}
\end{equation}
The probabilities used by this selector are generated by a reproducible \emph{patch probe}, not by TAP-Path itself. Specifically, Hibou-B patch embeddings are L2-normalized and a multinomial logistic-regression classifier is fitted on training patches only using the LBFGS solver, L2 penalty, $C=1.0$, and the fixed screening seed. For candidate patch $i$, if $\mathbf{p}_i^{(q)}\in[0,1]^{32}$ denotes the probe posterior, confidence and normalized predictive entropy are
\begin{align}
c_i &= \max_k p_{ik}^{(q)}, \label{eq:probeconf}\\
h_i &= -\frac{1}{\log C}\sum_{k=1}^{C}p_{ik}^{(q)}
\log\!\left(p_{ik}^{(q)}+\epsilon\right), \qquad C=32. \label{eq:probeentropy}
\end{align}
Thus, the probe is a linear probabilistic readout of frozen Hibou-B features; it is neither a ResNet nor a zero-shot classifier. It is trained exclusively on the training split and is used only to construct the deterministic patch manifest.

In Eq.~\eqref{eq:patchutility}, $c_i$ is the probe confidence from Eq.~\eqref{eq:probeconf}, $d_i$ is the minimum cosine distance between candidate $i$ and the already selected Hibou-B patch representations, and $h_i$ is the normalized entropy from Eq.~\eqref{eq:probeentropy}. The selection procedure first reserves one highest-confidence patch from every available scale, then forces the two highest-entropy remaining patches, and finally fills the bag according to Eq.~\eqref{eq:patchutility}. The diversity term prevents collapse to near-duplicate tissue regions, whereas the entropy term deliberately preserves difficult regions.

\begin{figure*}[!t]
\centering
\begin{subfigure}[t]{0.52\textwidth}
\centering
\includegraphics[width=\linewidth]{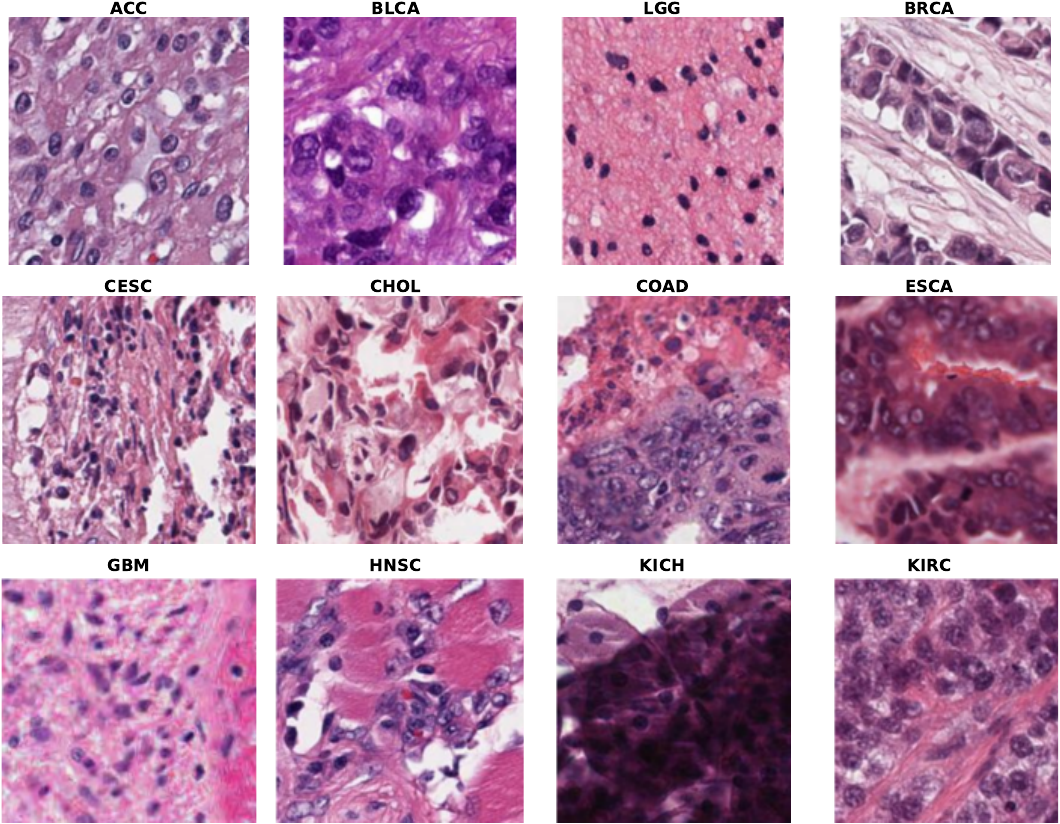}
\caption{Representative internal H\&E patches. TCGA abbreviations identify the sampled cancer classes.}
\end{subfigure}
\hfill
\begin{subfigure}[t]{0.46\textwidth}
\centering
\includegraphics[width=\linewidth]{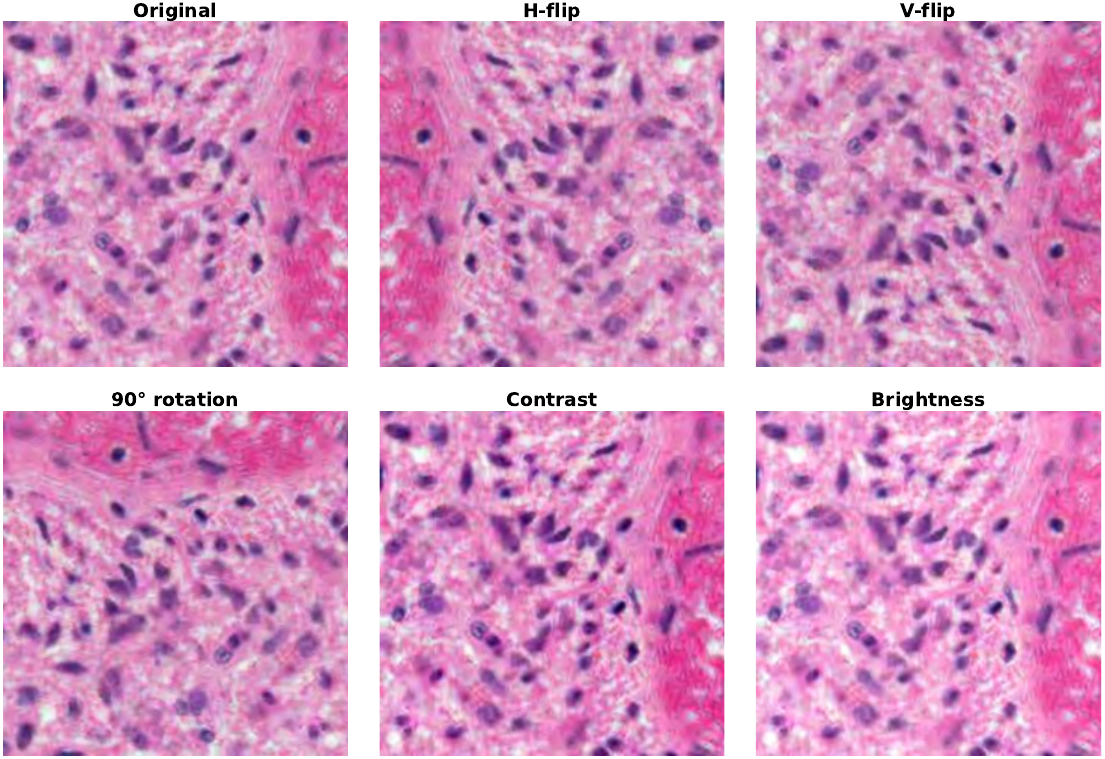}
\caption{Representative label-preserving transformations used during training.}
\end{subfigure}
\caption{Representative data appearance and training-time augmentation. The pathology panel illustrates morphological heterogeneity across selected internal classes, while the augmentation panel shows the conservative geometric and photometric perturbations used to improve robustness. Validation, test, and external evaluations use deterministic unaugmented patches.}
\label{fig:dataexamples}
\end{figure*}

\begin{figure}[t]
\centering
\includegraphics[width=\columnwidth]{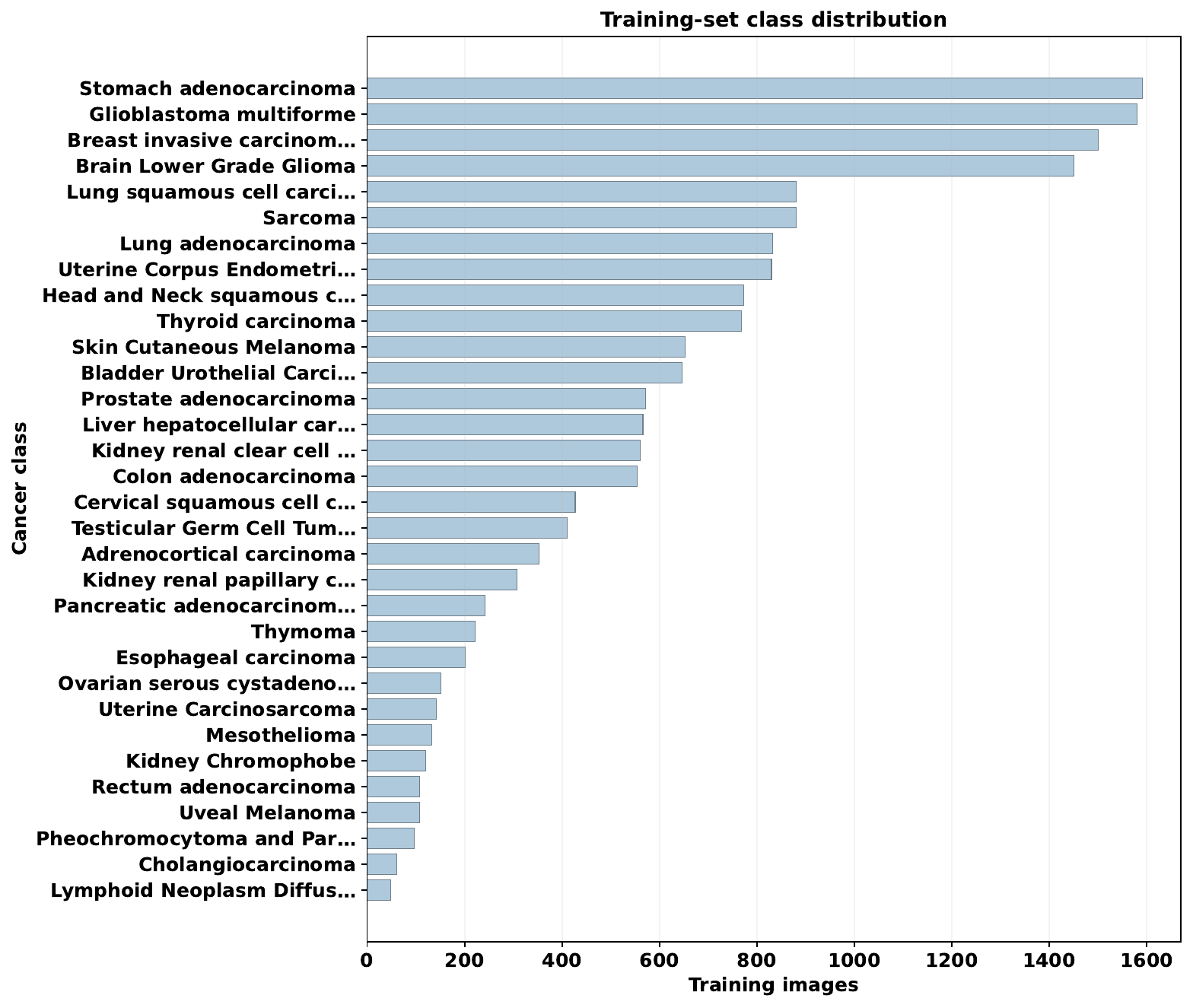}
\caption{Training-set distribution of the 32 internal cancer classes. The long-tailed class frequencies motivate balanced, macro-averaged, and rare-class evaluation.}
\label{fig:distribution}
\end{figure}

For source image $I$, stored crop center $(x_i,y_i)$, and source crop size $s_i$, the selected region is
\begin{equation}
P_i = I\left[
y_i-\frac{s_i}{2}:y_i+\frac{s_i}{2},
x_i-\frac{s_i}{2}:x_i+\frac{s_i}{2}
\right].
\end{equation}
Regions extending beyond the source boundary are padded with white background and resized to $224\times224$ pixels using bicubic interpolation. The resulting 12 regions are deterministic for a given manifest and are reused across compared encoders, preventing model-specific patch selection from contaminating the benchmark. Representative selected regions are shown in Fig.~\ref{fig:dataexamples}, while Fig.~\ref{fig:distribution} documents the long-tailed class distribution.

\FloatBarrier

Data-integrity checks verify the canonical class mapping, finite feature values, feature/index row agreement, deterministic split membership, and absence of zero-vector embeddings. As visualized in Fig.~\ref{fig:distribution}, class imbalance is substantial, motivating balanced accuracy, macro-F1, rare-class analysis, and class-wise reliability rather than relying exclusively on top-1 accuracy.

\subsection{Data augmentation}

Augmentation is restricted to transformations expected to preserve diagnostic class: horizontal/vertical reflection, $90^\circ$ rotation, mild contrast perturbation, and mild brightness perturbation. An augmented patch is
\begin{equation}
\tilde P = \mathcal{T}_{\theta}(P),\qquad
\mathcal{T}_{\theta}\in
\{\mathcal{F}_{h},\mathcal{F}_{v},\mathcal{R}_{90},
\mathcal{C}_{\delta_c},\mathcal{B}_{\delta_b}\}.
\end{equation}
Aggressive color remapping is avoided because stain-dependent appearance can carry diagnostic information. Validation, internal test, and external CPTAC evaluation use deterministic, unaugmented patches.

\subsection{Foundation-model baselines}

Four pathology foundation encoders were retained as interpretable reference points: Hibou-B, CONCH, \vir, and UNI2-h. Hibou-B and CONCH represent smaller pathology-specific encoders, while \vir and UNI2-h represent the high-capacity regime. Comparisons use the same 32-class task and dataset partitions from the established benchmark. We report each baseline's measured downstream predictive metrics together with the parameter count and a consistently defined analytical encoder-FLOP estimate. The proposed method is initialized from \vir because it provided a strong large-model baseline while exposing sufficient depth for structural analysis. For additional context, the study evaluated high-compute fusion systems. Virchow2+StaticTriFusion reached 87.43\% accuracy with 81.78\% macro-F1, and UNI2-h+DenseTriGate reached 87.91\% accuracy with 82.68\% macro-F1, but both execute three foundation representations simultaneously. Their approximate deployed backbone footprints are 808M/421.6G and 857M/532.4G (parameters/FLOPs), respectively, before small fusion-head overhead. These systems are therefore retained as high-compute context rather than direct deployment competitors.

\subsection{\model overview}

Let the pretrained encoder contain $L=32$ transformer blocks,
\begin{equation}
\mathcal{F}(X)=B_L\circ B_{L-1}\circ\cdots\circ B_1(E(X)),
\label{eq:parentencoder}
\end{equation}
where $E(\cdot)$ denotes patch embedding plus prefix-token construction and $B_\ell$ is transformer block $\ell$ in Eq.~\eqref{eq:parentencoder}. \model transforms $\mathcal{F}$ into a task-specific sparse encoder $\mathcal{F}_{\mathcal{S},\rho}$ parameterized by a retained block set $\mathcal{S}\subset\{1,\ldots,L\}$ and token-retention ratio $\rho$. The final model uses $|\mathcal{S}|=24$ and $\rho=0.70$ after the pruning point.

\begin{figure*}[t]
    \centering
    \includegraphics[width=\textwidth]{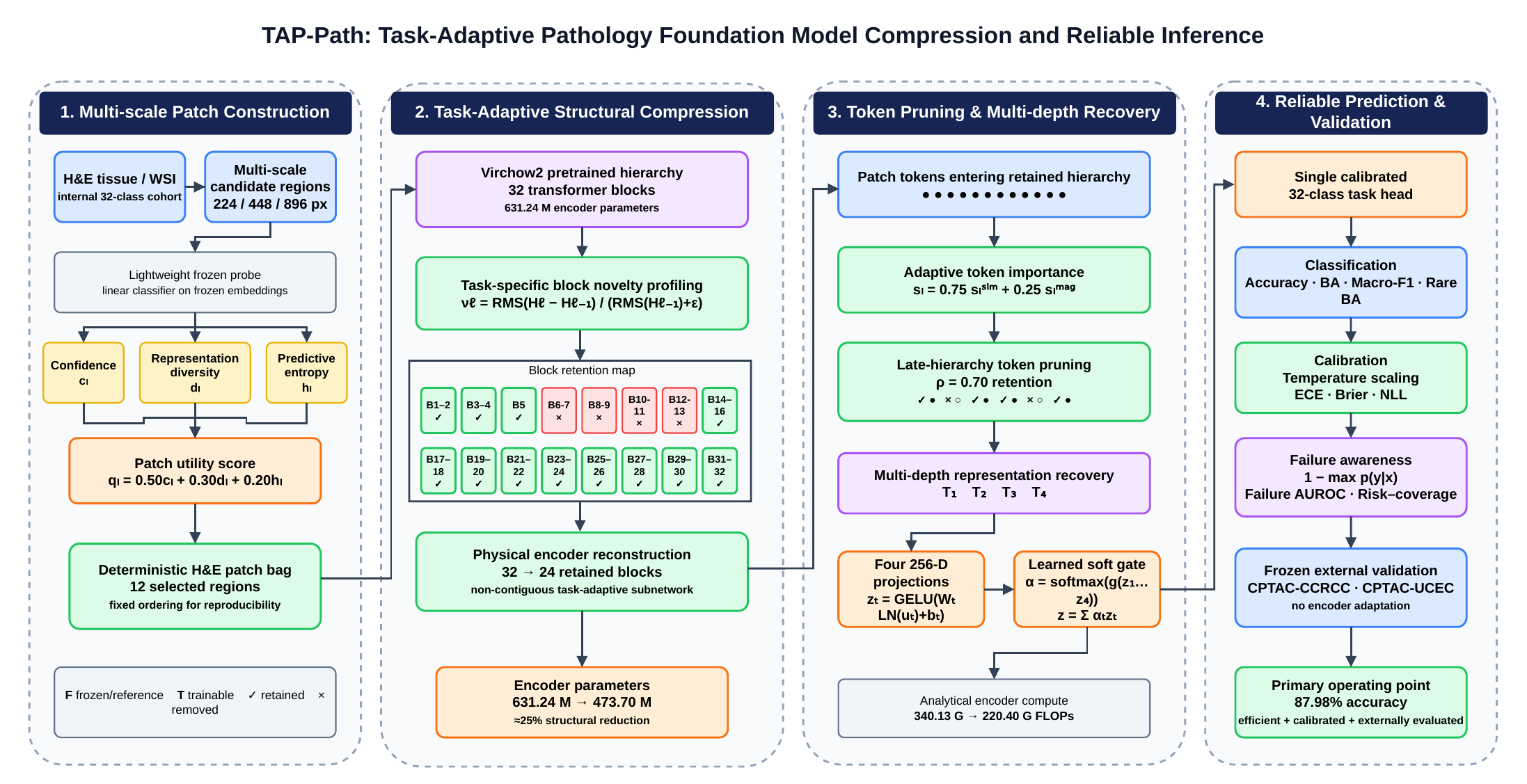}
    \caption{\textbf{Architecture of the proposed TAP-Path framework.}
    Multi-scale H\&E regions are ranked to construct a deterministic 12-patch bag, followed by task-adaptive structural compression of the Virchow2 encoder to the locked one-based block set $\{1,\ldots,5\}\cup\{14,\ldots,32\}$ (24 of 32 blocks). Adaptive late-hierarchy token pruning retains 70\% of informative tokens, while four intermediate feature representations are projected to 256 dimensions and adaptively combined through a learned soft gate before 32-class prediction. The resulting predictions are evaluated for classification performance, calibration, failure awareness, and frozen external generalization on CPTAC-CCRCC and CPTAC-UCEC. TAP-Path reduces the encoder from 631.24M to 473.70M parameters and the analytical encoder computation from 340.13 to 220.40 GFLOPs while achieving 87.98\% internal test accuracy.}
    \label{fig:architecture}
\end{figure*}

The design contains four coupled operations: (i) block novelty profiling and task-adaptive structural selection; (ii) physical removal of unselected transformer blocks; (iii) input-adaptive patch-token pruning; and (iv) gated recovery of features from four depths of the compressed hierarchy. The complete procedure is summarized in Algorithm~\ref{alg:tappath} and illustrated in Fig.~\ref{fig:architecture}; Table~\ref{tab:featurecompare} later contrasts the resulting system properties with the evaluated baselines.

\subsection{Task-adaptive block selection}

Uniform truncation assumes that later blocks are always more expendable or that useful information is distributed monotonically with depth. We instead estimate the normalized residual change induced by each block. For a profiling sample $x$ and block $\ell$, let $H_{\ell-1}$ and $H_\ell$ denote the token tensors immediately before and after the block. We define block novelty as
\begin{equation}
\nu_\ell(x)=
\frac{
\sqrt{\frac{1}{N D}\left\|H_\ell-H_{\ell-1}\right\|_F^2}
}{
\sqrt{\frac{1}{N D}\left\|H_{\ell-1}\right\|_F^2}+\epsilon
},
\label{eq:novelty}
\end{equation}
where $N$ is the token count, $D$ is the embedding dimension, and $\epsilon=10^{-8}$ prevents numerical instability. Profiling is restricted to the development subset and excludes internal test and CPTAC records. The dataset-level score is
\begin{equation}
\bar{\nu}_{\ell}=\frac{1}{M}\sum_{m=1}^{M}\nu_\ell(x_m).
\end{equation}

To preserve low-level token formation and high-level semantic consolidation, the first four and final four blocks are anchored. For a target retained depth $K$, the remaining $K-8$ blocks are selected from the middle hierarchy according to $\bar{\nu}_{\ell}$ and then restored to their original order. Because transformer blocks preserve a constant hidden width, block excision does not create a dimensional mismatch. If $H_{\ell}\in\mathbb{R}^{N\times D}$ is the residual stream after a retained block and blocks $\ell+1,\dots,j-1$ are removed, the next retained block receives
\begin{equation}
H_j=B_j(H_\ell).
\end{equation}
The pretrained weights inside retained modules are unchanged; only the computational graph is shortened. The selected 24-block configuration in the locked experiment retained original blocks
\begin{equation}
\mathcal{S}^{\star}=\{1,\ldots,5\}\cup\{14,\ldots,32\}.
\end{equation}
using one-based indexing for readability. The practical implementation physically replaces the original block list with the retained modules; consequently, omitted blocks no longer contribute stored encoder parameters or forward-pass computation.

Structural candidates were screened against contiguous-depth controls and task-sparse variants. Candidate selection was validation-only and explicitly constrained by compression. The objective can be written as
\begin{align}
\max_{\mathcal{S}} \quad & A_{\mathrm{val}}(\mathcal{S}) \\
\text{s.t.}\quad &
R_P(\mathcal{S})\ge r_P,\qquad
A_{\mathrm{val}}(\mathcal{S})\ge A^\star_{\mathrm{val}}-\delta,
\end{align}
where $R_P=1-P_{\mathcal{S}}/P_{\mathrm{full}}$ is parameter reduction, $r_P$ is the required reduction, and $\delta$ is the allowed validation tolerance. This formulation avoids choosing the smallest network when its accuracy has already collapsed.

\subsection{Adaptive token pruning}

After structural selection, token reduction is applied within the compressed encoder. Let $c\in\mathbb{R}^D$ denote the prefix/class token and $p_i\in\mathbb{R}^D$ a patch token. We first normalize both and compute a class-token similarity,
\begin{equation}
s_i^{\mathrm{sim}}=
\frac{p_i^\top c}{\|p_i\|_2\|c\|_2}.
\end{equation}
To avoid retaining only tokens aligned with the current class token, we add an activation-magnitude term. With per-image normalized magnitude
\begin{equation}
s_i^{\mathrm{mag}}=
\frac{\|p_i\|_2-\mu_{\|p\|}}{\sigma_{\|p\|}+\epsilon},
\end{equation}
the fixed development weighting prioritizes semantic alignment while retaining a secondary activation-magnitude cue; it is frozen before test evaluation. The combined token importance is
\begin{equation}
s_i=0.75\,s_i^{\mathrm{sim}}+0.25\,s_i^{\mathrm{mag}}.
\label{eq:token}
\end{equation}
The prefix token is always retained. Among $N_p$ patch tokens, the model keeps
\begin{equation}
K_p=\left\lceil \rho N_p \right\rceil.
\label{eq:tokenkeep}
\end{equation}
patches with the largest $s_i$ according to Eqs.~\eqref{eq:token}--\eqref{eq:tokenkeep}, where $\rho\in\{1.00,0.85,0.70\}$ was evaluated after structural locking. Token pruning begins at retained execution position $\max(1,\operatorname{round}(24\times0.55))=13$, i.e., after the 13th block in the retained execution sequence. Ranking is recomputed for every patch and introduces no trainable token-router subnetwork.

The analytical compute reduction arises from both fewer blocks and a smaller token sequence. If $\Phi_\ell(N)$ denotes the FLOPs of transformer block $\ell$ at token count $N$, the compressed encoder cost is approximated by
\begin{equation}
\Phi_{\mathrm{TAP}}=
\sum_{\ell\in\mathcal{S}_{\mathrm{pre}}}\Phi_\ell(N)
+
\sum_{\ell\in\mathcal{S}_{\mathrm{post}}}\Phi_\ell(K_p+N_{\mathrm{prefix}}),
\end{equation}
where $\mathcal{S}_{\mathrm{pre}}$ and $\mathcal{S}_{\mathrm{post}}$ denote retained blocks before and after the pruning point.

\subsection{Multi-depth feature recovery}

Compression can remove intermediate transformations that a downstream classifier would otherwise exploit. We therefore do not classify from only the final compressed token state. Four approximately evenly spaced taps are collected from the retained hierarchy. Four taps balance hierarchical coverage against projection/gating overhead and cached feature dimensionality. The effective token count $K_t$ depends on tap location: taps before the pruning point observe the dense sequence, whereas later taps observe the retained 70\% patch sequence. At tap $t$, the token sequence is normalized and summarized using the prefix token and mean patch token,
\begin{equation}
u_t=\left[\operatorname{LN}(c_t)\,;\,
\frac{1}{K_t}\sum_{i=1}^{K_t}\operatorname{LN}(p_{t,i})\right].
\end{equation}
Each $u_t$ is projected to a 256-dimensional task space:
\begin{equation}
z_t=\operatorname{GELU}(W_t \operatorname{LN}(u_t)+b_t).
\end{equation}

Instead of concatenating the four taps and treating them equally, an adaptive gate estimates their sample-specific contribution. Let
\begin{equation}
z_{\mathrm{cat}}=[z_1;z_2;z_3;z_4].
\end{equation}
The gate is
\begin{equation}
\bm{\alpha}=
\operatorname{softmax}
\left(
W_{g2}\,\operatorname{GELU}
\left(W_{g1}\operatorname{LN}(z_{\mathrm{cat}})\right)
\right),
\quad \sum_{t=1}^{4}\alpha_t=1,
\label{eq:gate}
\end{equation}
and the fused feature becomes
\begin{equation}
z=\sum_{t=1}^{4}\alpha_t z_t.
\label{eq:fusion}
\end{equation}
The final classifier is a two-layer MLP with LayerNorm, a 512-unit hidden layer, GELU, dropout 0.18, and a 32-class output layer. This single task head contains 5.70M parameters; the deployed single-head model therefore contains 479.40M parameters in total.


\subsection{Optimization objectives}

For the primary single-head model, class weights are mild rather than fully inverse-frequency weighted:
\begin{equation}
w_c=
\frac{
\left(\frac{N}{C\,n_c}\right)^\gamma
}{
\frac{1}{C}\sum_{j=1}^{C}
\left(\frac{N}{C\,n_j}\right)^\gamma
},
\qquad \gamma=0.20,
\end{equation}
where $n_c$ is the number of training examples in class $c$. Weighted cross-entropy uses label smoothing $\epsilon_{\mathrm{ls}}=0.01$. To discourage degenerate single-tap gating, we define gate entropy
\begin{equation}
H(\bm{\alpha})=-\sum_{t=1}^{4}\alpha_t\log(\alpha_t+\epsilon)
\end{equation}
and optimize
\begin{equation}
\mathcal{L}_{\mathrm{primary}}
=
\mathcal{L}_{\mathrm{CE}}
-
\lambda_H H(\bm{\alpha}),
\qquad \lambda_H=0.002.
\end{equation}

Rare-class sensitivity was investigated after the main architecture had been locked. Five validation-only objectives were compared: mild weighted cross-entropy, Balanced Softmax, logit adjustment with $\tau=0.5$, logit adjustment with $\tau=1.0$, and class-balanced focal loss. For logit adjustment,
\begin{equation}
\tilde{o}_c=o_c+\tau\log(\pi_c+\epsilon),
\end{equation}
where $o_c$ is class logit $c$ and $\pi_c$ is the empirical training prior. This experiment is reported as an ablation of the \emph{same compressed architecture}; it is not merged with the primary head to construct an artificial ``best of all metrics'' model.

\begin{algorithm*}[t]
\caption{TAP-Path: task-adaptive structural and token pruning with multi-depth recovery}
\label{alg:tappath}
\begin{algorithmic}[1]
\Require Pretrained Virchow2 encoder $\{B_\ell\}_{\ell=1}^{32}$; training set $\mathcal{D}_{tr}$; validation set $\mathcal{D}_{va}$; deterministic candidate-patch manifest
\Ensure Compressed encoder $\mathcal{F}_{\mathcal{S}^{\star},\rho^{\star}}$ and calibrated 32-class task head
\State Extract frozen Hibou-B candidate-patch embeddings and L2-normalize them
\State Fit the multinomial logistic patch probe on $\mathcal{D}_{tr}$ only
\For{each candidate patch $i$}
    \State Compute probe confidence $c_i$ and entropy $h_i$ using Eqs.~\eqref{eq:probeconf}--\eqref{eq:probeentropy}
\EndFor
\For{each image}
    \State Reserve the highest-confidence patch at every available scale
    \State Add the two highest-entropy remaining patches
    \While{fewer than 12 patches are selected}
        \State Compute confidence--diversity--entropy utility using Eq.~\eqref{eq:patchutility}
        \State Add the highest-utility remaining patch
    \EndWhile
\EndFor
\State Profile Virchow2 block novelty $\bar{\nu}_{\ell}$ on development data using Eq.~\eqref{eq:novelty}
\State Construct contiguous-depth controls and non-contiguous task-sparse block candidates
\For{each candidate block set $\mathcal{S}$}
    \State Physically instantiate only blocks in $\mathcal{S}$
    \State Extract reduced-protocol screening features and fit the fixed lightweight validation probe
    \State Record validation accuracy, BA, macro-F1, encoder parameters, and analytical FLOPs
\EndFor
\State Lock the validation-Pareto structural candidate $\mathcal{S}^{\star}$ (24 of 32 blocks)
\For{$\rho\in\{1.00,0.85,0.70\}$}
    \State Score tokens with Eq.~\eqref{eq:token}; retain $K_p$ tokens using Eq.~\eqref{eq:tokenkeep}
    \State Evaluate validation performance and analytical compute
\EndFor
\State Lock $\rho^{\star}=0.70$ and rebuild the executable compressed encoder
\For{seed $\in\{42,123,2026\}$}
    \State Extract the full deterministic 12-patch representation
    \State Recover four hierarchy taps and compute the adaptive gate using Eq.~\eqref{eq:gate}
    \State Fuse taps using Eq.~\eqref{eq:fusion} and optimize the single 32-class head
    \State Fit temperature scaling on validation logits only
    \State Evaluate the locked internal test set and frozen CPTAC cohorts
\EndFor
\State Report predictive, efficiency, calibration, failure-detection, selective-risk, rare-class, and bootstrap metrics
\end{algorithmic}
\end{algorithm*}

\subsection{Training and model selection}
\label{sec:training}

The optimization pipeline is separated into architecture selection and task-head learning so that the expensive foundation backbone is not repeatedly fine-tuned for every candidate.

\paragraph{Stage 1: reduced class-aware screening.}
Structural search uses 1,400 training and 700 validation records selected deterministically with class awareness. Only the first two ranked patches per selected image are processed during screening. This subset exists solely to rank compression candidates efficiently.

\paragraph{Stage 2: structural search.}
Contiguous-depth controls and non-contiguous task-sparse candidates are constructed from the frozen \vir parent. Each candidate is evaluated using screening features and a lightweight validation probe. The selection criterion jointly considers validation utility and compression; internal test data are not used to choose retained blocks.

\paragraph{Stage 3: token-retention search.}
After the strongest compressed structures are identified, $\rho\in\{1.00,0.85,0.70\}$ is evaluated on validation data. TaskSparse24 with $\rho=0.70$ is locked before final representation extraction.

\paragraph{Stage 4: full 12-patch feature extraction.}
The physically reconstructed encoder is run over the complete deterministic 12-patch bags. CUDA automatic mixed precision is enabled where supported. Screening encoder batch size is 4, final feature-extraction batch size is 6, and the development GPU is an NVIDIA GeForce RTX 5060 Ti with approximately 16 GB VRAM.

\paragraph{Stage 5: task-head optimization.}
The single gated multi-tap head is trained on cached image-level compressed features for at most 130 epochs using AdamW, learning rate $7\times10^{-4}$, weight decay $2\times10^{-4}$, batch size 384, gradient clipping at 5, dropout 0.18, and early-stopping patience 18. Three seeds (42, 123, 2026) quantify optimization variability.

\paragraph{Stage 6: calibration and locked evaluation.}
Temperature scaling is fitted only to validation logits using LBFGS. The locked internal test set is then evaluated. CPTAC-CCRCC and CPTAC-UCEC are processed only after internal configuration locking; no external label, threshold, temperature, or architecture decision modifies TAP-Path.

Structural-screening performance and final test performance are not directly comparable because they correspond to distinct experimental protocols. Screening uses a reduced class-aware subset and two patches per image for efficient candidate ranking, whereas final evaluation uses the complete deterministic 12-patch representation and the trained multi-depth recovery head.

\subsection{Common-probe diagnostic}

To assess representation quality independently of model-specific task heads, we additionally trained an identical lightweight probe on each frozen representation. Under this standardized readout, TAP-Path achieved approximately 85\% accuracy, compared with approximately 88\% using its proposed multi-depth gated head. The difference reflects the contribution of the task-specific multi-depth recovery mechanism to downstream prediction. Because the common-probe experiment evaluates representation quality under a standardized classifier rather than the complete TAP-Path inference architecture, it is reported as a complementary diagnostic analysis.

\subsection{Reliability metrics}

For each trained seed, post-hoc temperature scaling is fitted on validation logits only. Given logits $\bm{o}$ and temperature $T>0$,
\begin{equation}
p_c=
\frac{\exp(o_c/T)}
{\sum_{j=1}^{C}\exp(o_j/T)}.
\end{equation}
$T$ is optimized by minimizing validation NLL. ECE is computed over 15 confidence bins,
\begin{equation}
\ece=
\sum_{b=1}^{B}\frac{|I_b|}{n}
\left|
\operatorname{acc}(I_b)-\operatorname{conf}(I_b)
\right|.
\end{equation}
We additionally report multiclass Brier score
\begin{equation}
\operatorname{Brier}=
\frac{1}{n}\sum_{i=1}^{n}
\sum_{c=1}^{C}
(p_{ic}-y_{ic})^2
\end{equation}
and NLL. These measures are complementary: ECE summarizes bin-level calibration, while NLL and Brier are proper scoring rules sensitive to the full predictive distribution.

For failure detection, each example receives an uncertainty score
\begin{equation}
u_i=1-\max_c p_{ic}.
\end{equation}
Ground-truth failures are $e_i=\mathbb{1}[\arg\max_c p_{ic}\neq y_i]$. The AUROC between $u_i$ and $e_i$ measures whether errors tend to receive higher uncertainty. Risk--coverage analysis sorts examples by confidence and evaluates error rate among the retained fraction. This supports a selective-prediction interpretation but is not presented as proof of clinical safety.

Rare classes are defined from the training distribution using the lower quartile of positive class counts. Rare-class balanced accuracy is the unweighted mean recall over those classes:
\begin{equation}
\mathrm{RareBA}=
\frac{1}{|\mathcal{R}|}
\sum_{c\in\mathcal{R}}
\frac{\mathrm{TP}_c}{\mathrm{TP}_c+\mathrm{FN}_c}.
\end{equation}

\subsection{Statistical analysis}
Three independent seeds are reported as mean$\pm$SD. Stratified bootstrap confidence intervals are estimated from frozen test predictions using $B=2000$ resamples:
\begin{equation}
\mathrm{CI}_{95}(m)=
\left[Q_{0.025}\{m_b\}_{b=1}^{B},
Q_{0.975}\{m_b\}_{b=1}^{B}\right].
\end{equation}
Because aligned prediction vectors are not available for every baseline, marginal confidence intervals are not used as a substitute for a paired superiority test. Small accuracy differences are therefore described as comparable/slightly higher performance, whereas parameter and FLOP reductions are reported as deterministic architectural differences.

\subsection{Implementation details}

Experiments were implemented in Python 3.11 with PyTorch and the \texttt{timm} model library. Foundation models were used as pretrained encoders; mixed-precision feature extraction was enabled on CUDA where supported. The reported development system used an NVIDIA GeForce RTX 5060 Ti GPU. Structural screening used a reduced validation protocol to avoid repeatedly executing all 12 patches for every candidate, whereas the locked winner was re-evaluated under the full 12-patch protocol.

The final single-head task model used four taps, 256-dimensional tap projections, dropout 0.18, up to 130 training epochs, early stopping with patience 18, AdamW with learning rate $7\times10^{-4}$ and weight decay $2\times10^{-4}$, gradient clipping at 5, and three seeds (42, 123, 2026). The task-head batch size was 384 on cached image-level representations. Temperature scaling used LBFGS on the validation logits. The final runtime profile of the compressed model was 31.85$\pm$1.21 ms per image on the development GPU, corresponding to 31.40 images/s under the measured profiling configuration.

\begin{table*}[t]
\centering
\caption{Method-level comparison of the evaluated foundation-model strategies. A checkmark indicates that the property is explicitly present in the evaluated system.}
\label{tab:featurecompare}
\small
\resizebox{\textwidth}{!}{%
\begin{tabular}{lccccc}
\toprule
Method & Physical compression & Adaptive token pruning & Multi-depth recovery & Reliability analysis & Frozen external test\\
\midrule
Hibou-B & -- & -- & -- & -- & \cmark\\
CONCH & -- & -- & -- & -- & \cmark\\
Virchow2 & -- & -- & -- & -- & \cmark\\
UNI2-h & -- & -- & -- & -- & \cmark\\
StaticTriFusion / DenseTriGate & -- & -- & \cmark & partial & \cmark\\
\textbf{TAP-Path} & \textbf{\cmark} & \textbf{\cmark} & \textbf{\cmark} & \textbf{\cmark} & \textbf{\cmark}\\
\bottomrule
\end{tabular}%
}
\end{table*}

\section{Results}

\subsection{Structural compression}

The first question was whether selecting blocks by task-dependent novelty provides a better compressed representation than simply truncating the model. Table~\ref{tab:struct} summarizes the validation-only screening. Full Depth32 achieved the highest absolute screening accuracy (72.96\%) but retained the entire encoder. A contiguous Depth24 model reduced encoder parameters by 24.96\% but reached 68.28\% screening accuracy. At the same parameter count, TaskSparse24 improved screening accuracy to 69.79\%, supporting the use of non-contiguous task-adaptive block retention. More aggressive TaskSparse22 and TaskSparse20 configurations reduced parameters by 31.20\% and 37.43\%, respectively, but incurred additional validation loss.

\begin{table}[t]
\centering
\caption{Structural validation screen. Values are used only for architecture selection and are not the final 12-patch test performance.}
\label{tab:struct}
\small
\begin{tabular}{lrrrr}
\toprule
Candidate & Acc. & BA & Params & FLOPs\\
 & (\%) & (\%) & (M) & (G)\\
\midrule
Depth32 & 72.96 & 73.08 & 631.24 & 340.13\\
Depth28 & 70.39 & 70.31 & 552.47 & 297.66\\
Depth24 & 68.28 & 68.38 & 473.70 & 255.19\\
TaskSparse24 & 69.79 & 69.84 & 473.70 & 255.19\\
TaskSparse22 & 69.18 & 69.30 & 434.32 & 233.96\\
TaskSparse20 & 66.47 & 66.50 & 394.94 & 212.73\\
\bottomrule
\end{tabular}
\end{table}

\begin{figure}[t]
\centering
\includegraphics[width=\columnwidth]{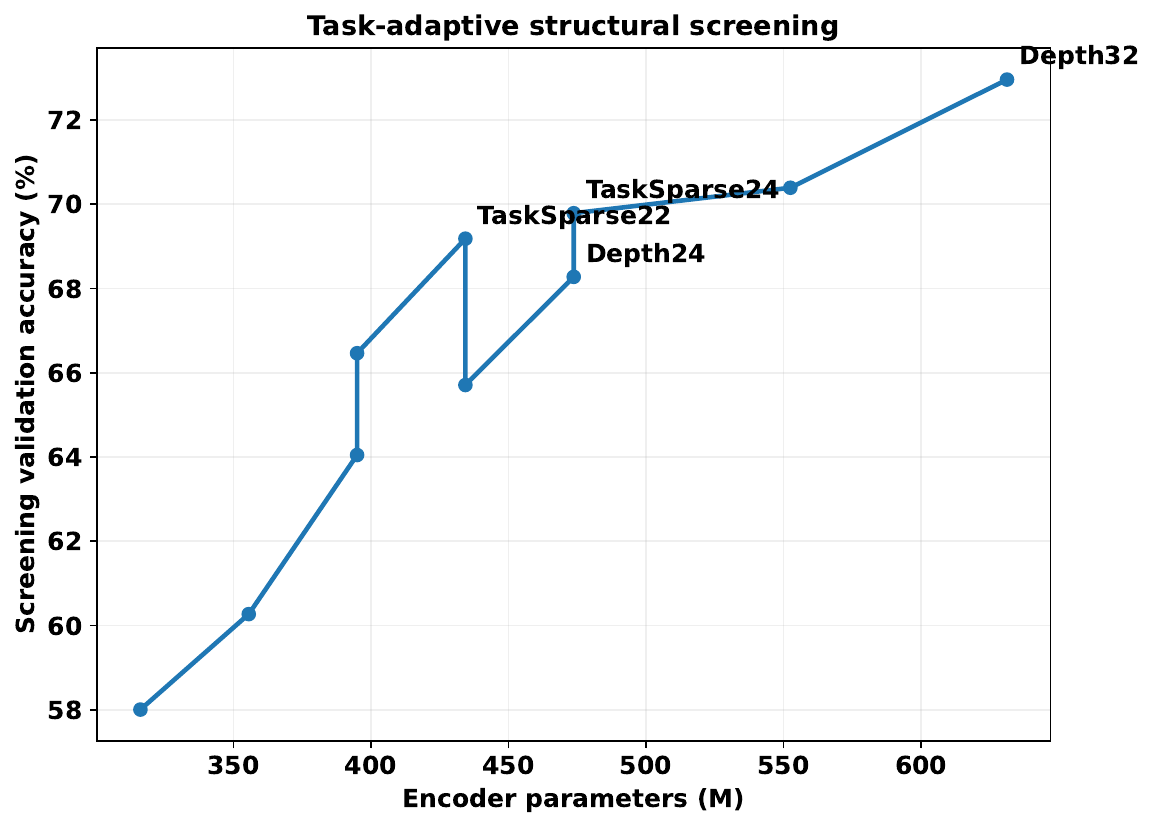}
\caption{Validation-only structural screening of contiguous-depth and task-adaptive sparse candidates.}
\label{fig:structscreen}
\end{figure}

Figure~\ref{fig:structscreen} visualizes the screening frontier. Importantly, the screening accuracies are not directly compared with the final test accuracies because they were obtained with a reduced screening protocol. Their purpose was model selection under equal candidate conditions.

\subsection{Token-pruning ablation}

\begin{figure}[t]
\centering
\includegraphics[width=\columnwidth]{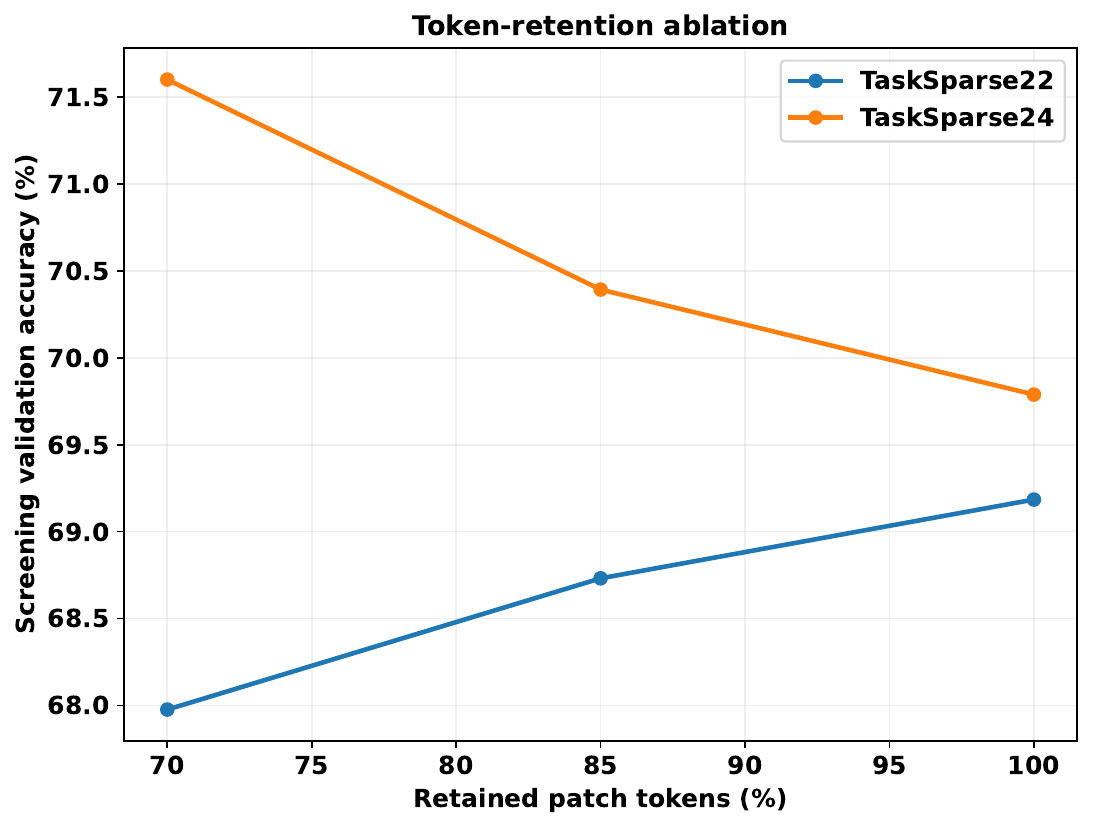}
\caption{Token-retention ablation after structural selection. The 70\% TaskSparse24 configuration provides the best validation accuracy with the lowest compute among the displayed TaskSparse24 settings.}
\label{fig:token}
\end{figure}

Token-retention results are shown in Table~\ref{tab:token}, and the corresponding validation trend is visualized in Fig.~\ref{fig:token}. For TaskSparse24, retaining 85\% of patch tokens increased validation accuracy from 69.79\% to 70.39\% while reducing analytical FLOPs from 255.19G to 237.71G. Retaining 70\% produced the strongest validation score (71.60\%) and further reduced compute to 220.40G, a 35.20\% reduction relative to full \vir. The same 70\% ratio was less effective for TaskSparse22, indicating an interaction between structural depth and token sparsity rather than a universal benefit from dropping tokens.

\begin{table}[t]
\centering
\caption{Token-retention ablation for the locked structural candidates.}
\label{tab:token}
\small
\begin{tabular}{lrrrr}
\toprule
Structure & Tokens & Acc. & BA & FLOPs\\
 & (\%) & (\%) & (\%) & (G)\\
\midrule
TaskSparse24 & 100 & 69.79 & 69.84 & 255.19\\
TaskSparse24 & 85 & 70.39 & 70.42 & 237.71\\
\textbf{TaskSparse24} & \textbf{70} & \textbf{71.60} & \textbf{71.64} & \textbf{220.40}\\
TaskSparse22 & 100 & 69.18 & 69.30 & 233.96\\
TaskSparse22 & 85 & 68.73 & 68.56 & 218.07\\
TaskSparse22 & 70 & 67.98 & 67.97 & 202.33\\
\bottomrule
\end{tabular}
\end{table}

The validation results indicate that moderate adaptive token pruning can improve the accuracy--compute operating point after stable early representations have formed. This behavior is consistent with removal of redundant or weakly aligned token content and motivates joint selection of structural depth and token retention under the validation protocol.

\subsection{Foundation-model comparison}

Table~\ref{tab:main} presents the central same-task comparison. The proposed single-head \model configuration achieved $87.98\pm0.067$\% test accuracy across three seeds, compared with 86.89\% for full \vir and 87.67\% for UNI2-h. The corresponding macro-F1 was $82.38\pm0.48$\%, compared with 80.94\% for \vir and 81.75\% for UNI2-h, while balanced accuracy reached $81.26\pm0.49$\%. These results place \model at the highest observed accuracy and macro-F1 among the evaluated single-backbone systems while substantially reducing the computational footprint of the large-model parent. Because paired baseline prediction files were unavailable, the comparison is reported as an observed performance difference without a paired significance test.

The efficiency gains are substantial. Full \vir contains approximately 631--632M encoder parameters, whereas the physically compressed encoder contains 473.70M, corresponding to a 24.96\% reduction. Including the task head, the deployed \model configuration contains 479.40M parameters, and analytical encoder FLOPs decrease from 340.13G to 220.40G (35.20\%). Relative to UNI2-h, \model also operates with substantially fewer parameters and analytical FLOPs while achieving comparable or higher predictive performance. Hibou-B and CONCH occupy a lower-compute regime, whereas \model defines a favorable operating point within the high-performing large-foundation-model regime.

\begin{table*}[t]
\centering
\caption{Direct same-task comparison on the 32-class internal test set. \model values are mean$\pm$SD over three independently trained single heads. Arrows indicate preferred direction. Analytical FLOPs refer to the encoder and are used for relative efficiency comparison.}
\label{tab:main}
\small
\begin{tabular}{lrrrrrr}
\toprule
Model & Params (M) $\downarrow$ & FLOPs (G) $\downarrow$ & Accuracy (\%) $\uparrow$ & BA (\%) $\uparrow$ & Macro-F1 (\%) $\uparrow$ & External Acc. (\%) $\uparrow$\\
\midrule
Hibou-B & 85.7 & 46.32 & 82.67 & 75.17 & 76.33 & \textbf{91.53}\\
CONCH & 90.0 & 35.13 & 81.43 & 76.39 & 76.11 & 85.76\\
Virchow2 & 632.0 & 340.13 & 86.89 & 80.52 & 80.94 & 90.76\\
UNI2-h & 681.0 & 450.97 & 87.67 & 81.12 & 81.75 & 90.07\\
\textbf{\model (ours)} & \textbf{479.40} & \textbf{220.40} & \textbf{87.98$\pm$0.067} & \textbf{81.26$\pm$0.49} & \textbf{82.38$\pm$0.48} & 91.22$\pm$0.83\\
\bottomrule
\end{tabular}
\begin{flushleft}\footnotesize
External accuracy is reported alongside the internal metrics to characterize transfer behavior across model scales; Hibou-B attains the highest raw accuracy on the two-class CPTAC cohort, while \model provides the strongest joint internal accuracy--efficiency operating point among the high-capacity systems.
\end{flushleft}
\end{table*}

\begin{table*}[t]
\centering
\caption{Broader experimental context including high-compute fusion references. Fusion parameter/FLOP values approximate the simultaneously deployed foundation encoders and therefore indicate deployment scale rather than trainable task-head size.}
\label{tab:broader}
\small
\begin{tabular}{lrrrrl}
\toprule
System & Params (M) & FLOPs (G) & Accuracy (\%) & Macro-F1 (\%) & Deployment regime\\
\midrule
Virchow2 & 632.0 & 340.1 & 86.89 & 80.94 & Full single backbone\\
UNI2-h & 681.0 & 451.0 & 87.67 & 81.75 & Full single backbone\\
Virchow2+StaticTriFusion & $\sim$808 & $\sim$421.6 & 87.43 & 81.78 & Three-FM fusion\\
UNI2-h+DenseTriGate & $\sim$857 & $\sim$532.4 & 87.91 & \textbf{82.68} & Three-FM fusion\\
\textbf{TAP-Path} & \textbf{479.40} & \textbf{220.4} & \textbf{87.98} & 82.38 & \textbf{Compressed single backbone}\\
\bottomrule
\end{tabular}
\end{table*}

Table~\ref{tab:broader} places the high-compute fusion experiments beside the single-backbone systems. These experiments show that comparable discrimination can also be obtained through multi-foundation-model fusion, but at substantially greater deployment-scale parameter and compute requirements. In contrast, TAP-Path achieves its operating point with a single physically compressed backbone.

\begin{figure}[t]
\centering
\includegraphics[width=\columnwidth]{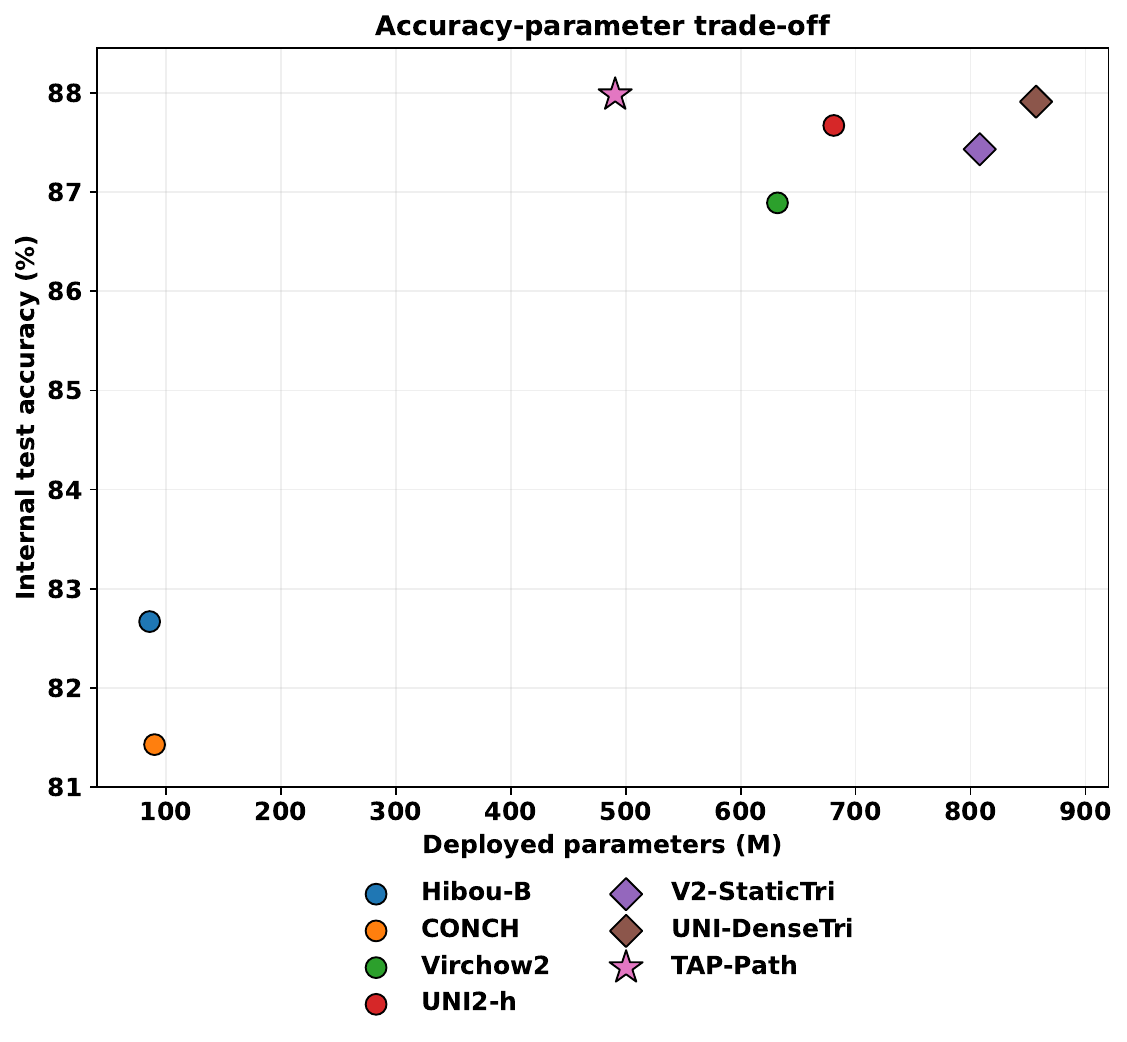}
\caption{Accuracy--parameter trade-off. Each operating point is labeled directly with model name and test accuracy; light guide lines preserve point--annotation correspondence.}
\label{fig:parampareto}
\end{figure}

\begin{figure}[t]
\centering
\includegraphics[width=\columnwidth]{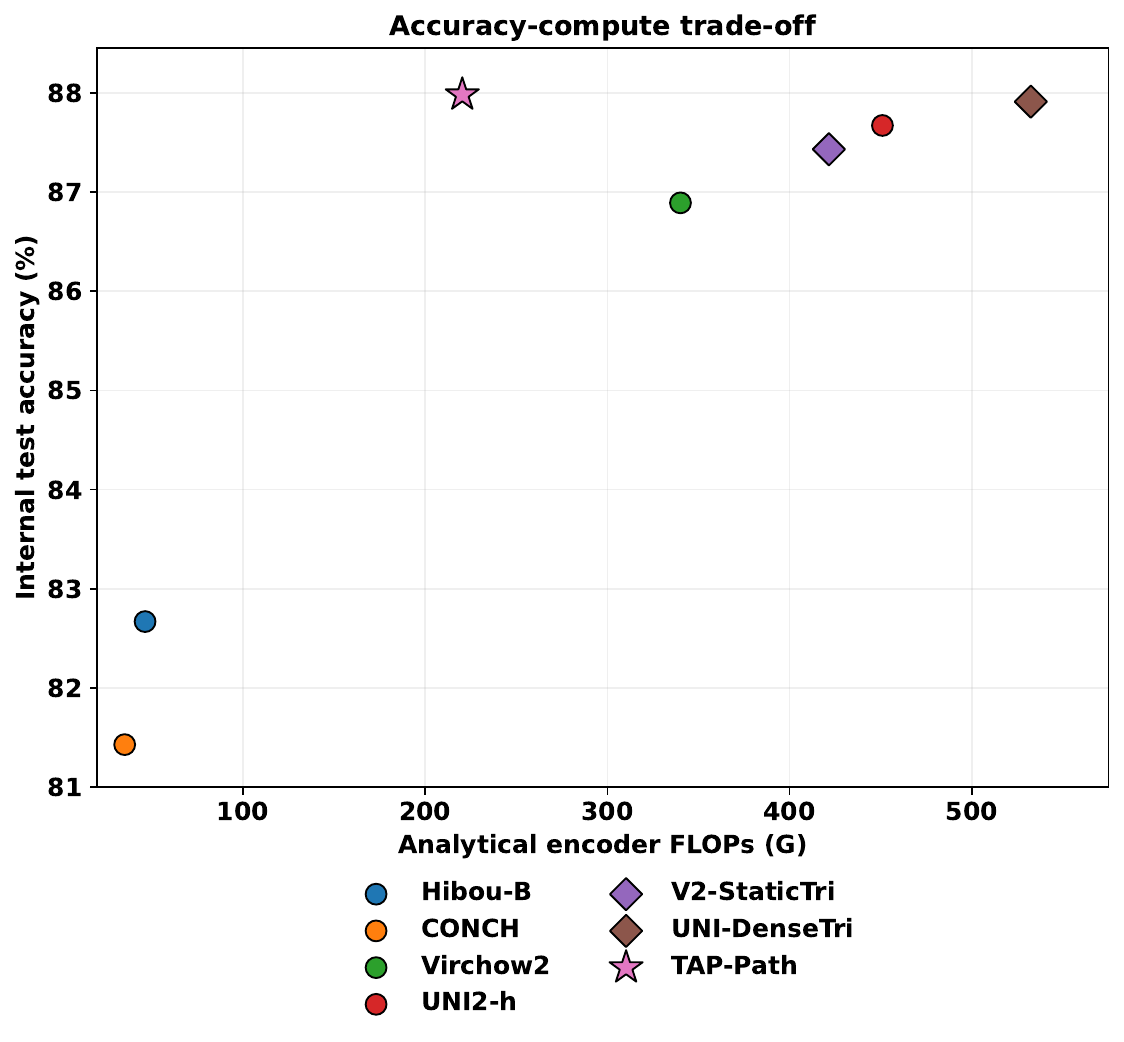}
\caption{Accuracy--compute trade-off. Direct annotations connect each model and test accuracy to its analytical encoder FLOP requirement.}
\label{fig:floppareto}
\end{figure}

The Pareto view is clearer in Figs.~\ref{fig:parampareto} and \ref{fig:floppareto}. The corrected high-compute fusion references are included in both plots: Virchow2+StaticTriFusion reaches 87.43\% accuracy and UNI2-h+DenseTriGate reaches 87.91\%, whereas TAP-Path reaches 87.98\%. Including the high-compute fusion references further illustrates the resulting Pareto trade-off. TAP-Path achieves 87.98\% accuracy using 479.40M deployed parameters and 220.40G analytical encoder FLOPs, while the fusion systems require substantially greater deployment-scale parameters and compute for similar predictive performance.

\subsection{Reliability analysis}
\label{sec:reliability}

\begin{figure}[t]
\centering
\includegraphics[width=\columnwidth]{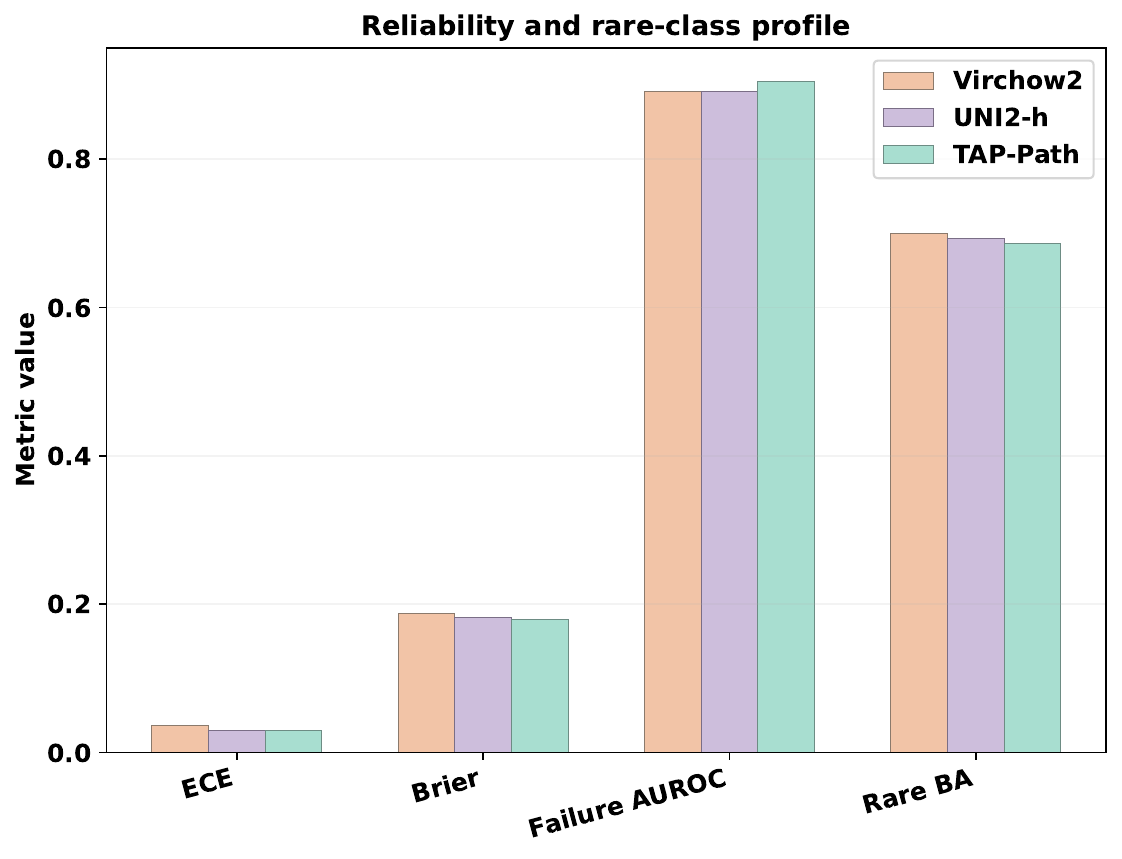}
\caption{Reliability and rare-class profile of TAP-Path relative to full Virchow2 and UNI2-h. Lower ECE/Brier and higher failure AUROC/Rare BA are preferred.}
\label{fig:trustprofile}
\end{figure}

\begin{figure}[t]
\centering
\includegraphics[width=\columnwidth]{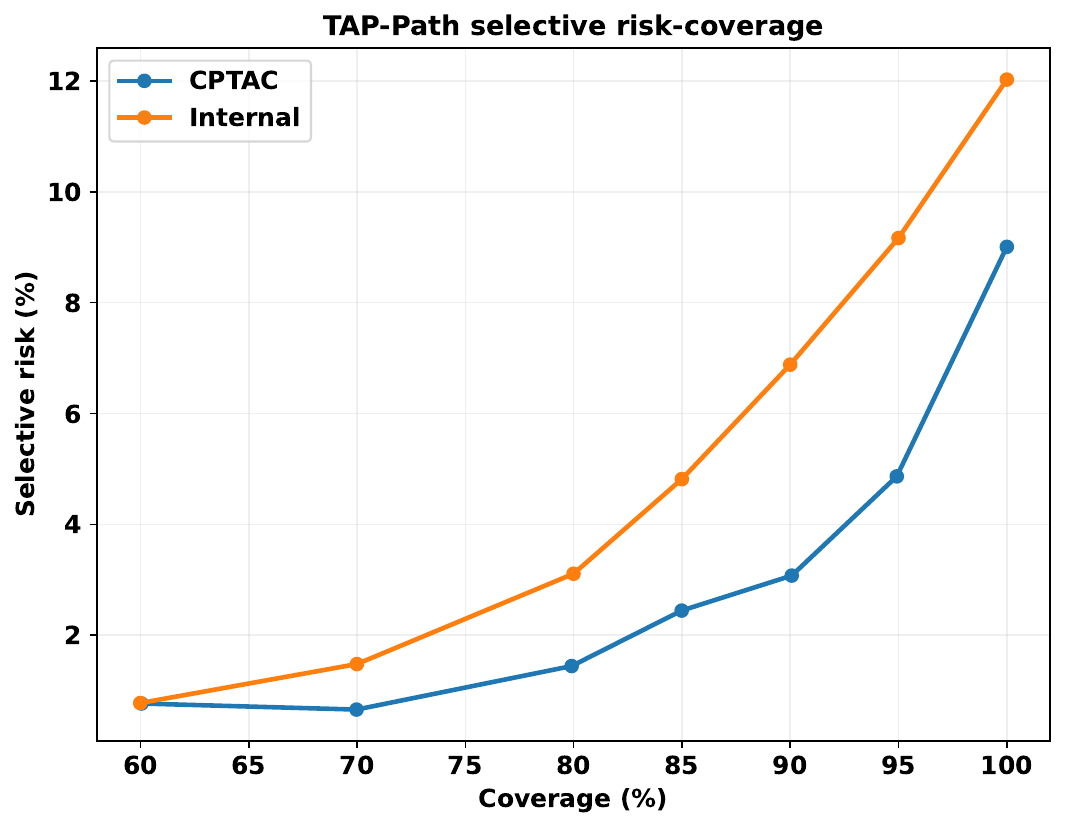}
\caption{Selective risk--coverage behavior for TAP-Path on the locked internal and external evaluations.}
\label{fig:riskcoverage}
\end{figure}

Reliability is organized into three complementary questions. First, \emph{are predicted probabilities well behaved?} This is assessed with ECE, NLL, and Brier score. Second, \emph{does confidence identify likely mistakes?} This is assessed with failure-detection AUROC and selective risk--coverage. Third, \emph{does performance remain acceptable for low-frequency classes?} This is assessed with rare-class balanced accuracy and class-wise F1. Table~\ref{tab:trust} and Figs.~\ref{fig:trustprofile}--\ref{fig:riskcoverage} report these three views jointly. For the primary three-seed TAP-Path task head, ECE is $0.0301\pm0.0022$, NLL is $0.4694\pm0.0015$, Brier score is $0.1800\pm0.0005$, and failure-detection AUROC is $0.9047\pm0.0060$. These measurements indicate that TAP-Path maintains informative confidence estimates despite structural and token pruning. In particular, its Brier score and failure-detection AUROC improve over the full Virchow2 baseline, whereas UNI2-h remains marginally better in ECE and NLL. The reliability claim is therefore based on the joint behavior of several metrics rather than on a single calibration statistic.

Rare-class sensitivity is evaluated on the same compressed encoder using the validation-selected rare-aware objective. This operating point achieves rare-class balanced accuracy $0.6864\pm0.0216$ and balanced accuracy $0.8229\pm0.0038$. We report that value in the reliability comparison because it is the dedicated minority-sensitive operating point of TAP-Path; the remaining TAP-Path entries in Table~\ref{tab:trust} are the primary three-seed reliability results. This distinction avoids understating the model's validated rare-class capability while retaining the provenance of each metric.

\begin{table}[t]
\centering
\caption{Reliability comparison for the large-model baselines and TAP-Path. TAP-Path ECE, Brier, and failure AUROC and Rare BA are the primary three-seed values at the operating point on the same compressed encoder.}
\label{tab:trust}
\scriptsize
\resizebox{\columnwidth}{!}{%
\begin{tabular}{lrrrr}
\toprule
Model & ECE $\downarrow$ & Brier $\downarrow$ & Fail-AUROC $\uparrow$ & Rare BA $\uparrow$\\
\midrule
Virchow2 & 0.0368 & 0.1882 & 0.8920 & \textbf{0.6996}\\
UNI2-h & \textbf{0.0297} & 0.1825 & 0.8920 & 0.6931\\
\textbf{TAP-Path} & 0.0301$\pm$.0022 & \textbf{0.1800$\pm$.0005} & \textbf{0.9047$\pm$.0060} & 0.6864$\pm$.0216\\
\bottomrule
\end{tabular}%
}
\end{table}

Figure~\ref{fig:trustprofile} provides a direct visual comparison of the reliability profile, while Fig.~\ref{fig:riskcoverage} evaluates whether confidence can support selective prediction. At 60\% internal coverage, TAP-Path retains approximately 99.2\% selective accuracy, and risk rises gradually as progressively less-confident cases are accepted. This monotonic behavior is consistent with a useful abstention signal: the model's low-confidence subset is enriched for errors.

\subsection{Rare-class analysis}

Following the reliability comparison in Table~\ref{tab:trust}. Rare-class behavior was further examined under alternative optimization objectives while keeping the compressed TAP-Path encoder fixed. The primary calibrated configuration is denoted by P, whereas the validation-selected rare-aware configuration is denoted by R. This separation allows the effect of the task-head objective on minority-class sensitivity to be evaluated independently of structural compression.


The rare-aware objectives are therefore reported within the unified ablation analysis in Table~\ref{tab:ablation}, preserving the distinction between the primary operating point and alternative minority-sensitive objectives. Instead, Table~\ref{tab:ablation} integrates the rare-aware objectives into the full ablation narrative alongside structural and token choices. On validation data, Balanced Softmax produced the highest rare-class balanced accuracy (70.55\%) but reduced overall accuracy. Logit adjustment with $\tau=1.0$ achieved the highest validation accuracy (90.25\%) among the tested objectives while raising rare-class BA from 67.39\% for the mild-CE baseline to 69.68\%.

\begin{figure}[t]
\centering
\includegraphics[width=\columnwidth]{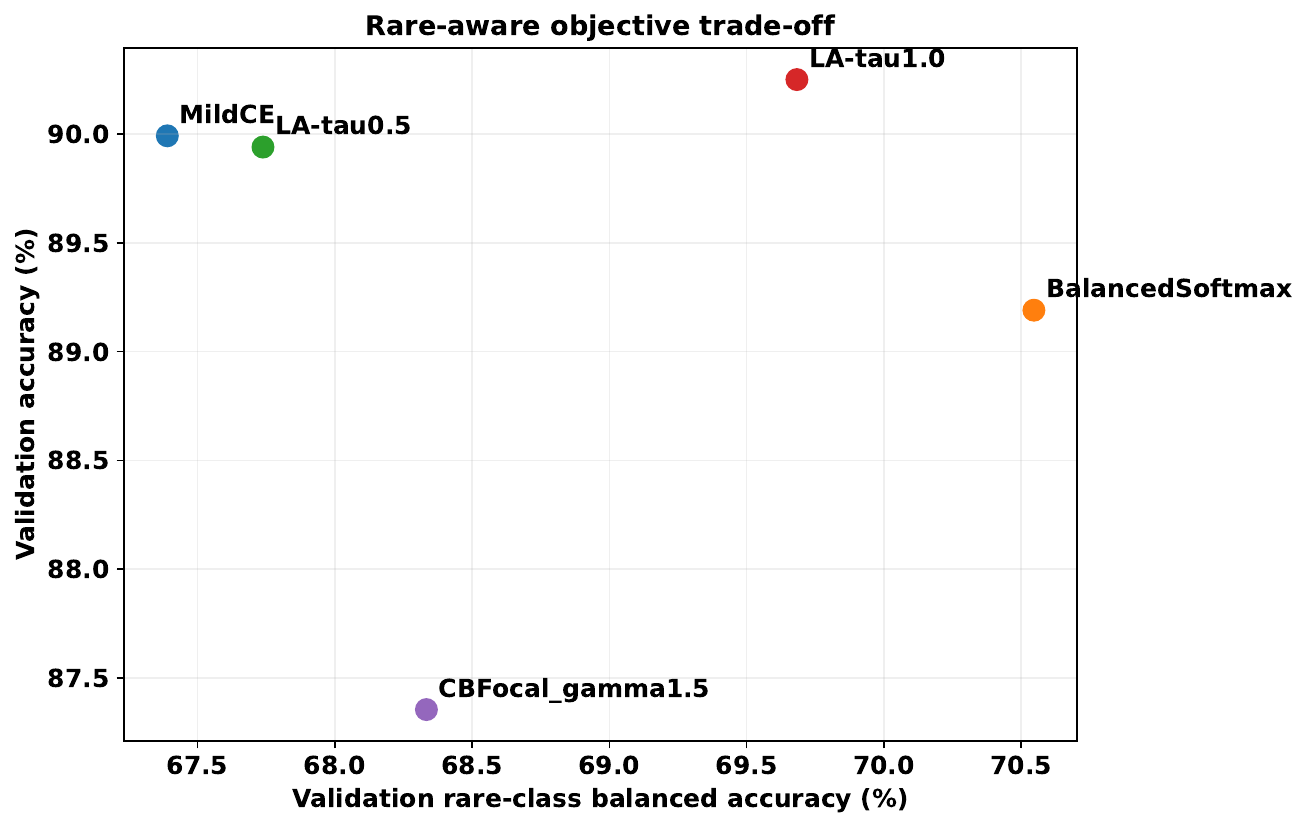}
\caption{Validation trade-off between overall accuracy and rare-class balanced accuracy for alternative task-head objectives on the fixed TAP-Path encoder.}
\label{fig:rareloss}
\end{figure}

\begin{figure*}[t]
\centering
\begin{subfigure}[t]{0.40\textwidth}
\centering
\includegraphics[height=7.0cm,keepaspectratio]{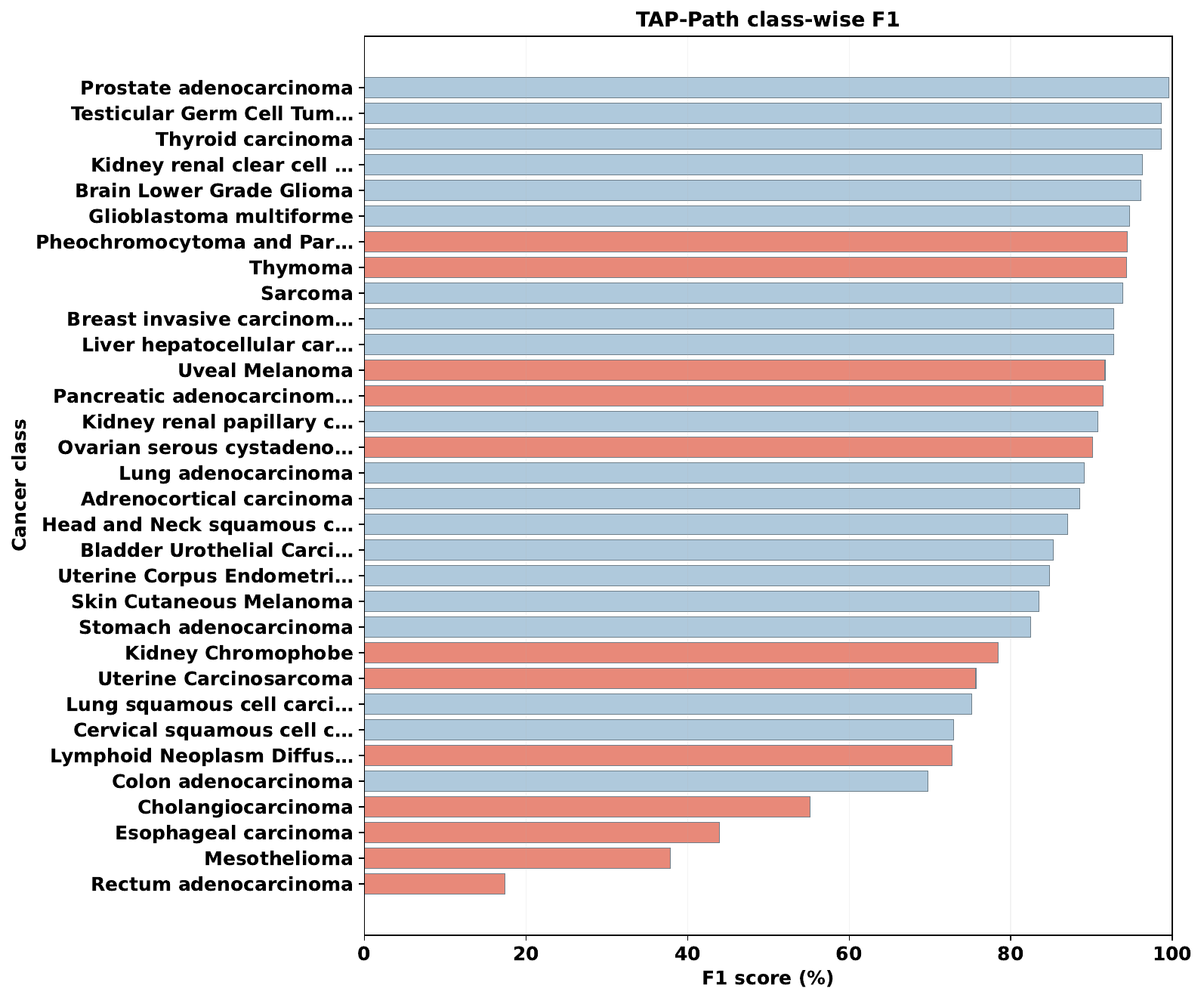}
\caption{Per-class F1. Rare classes are highlighted with the contrasting bar color.}
\end{subfigure}
\hfill
\begin{subfigure}[t]{0.57\textwidth}
\centering
\includegraphics[height=7.0cm,keepaspectratio]{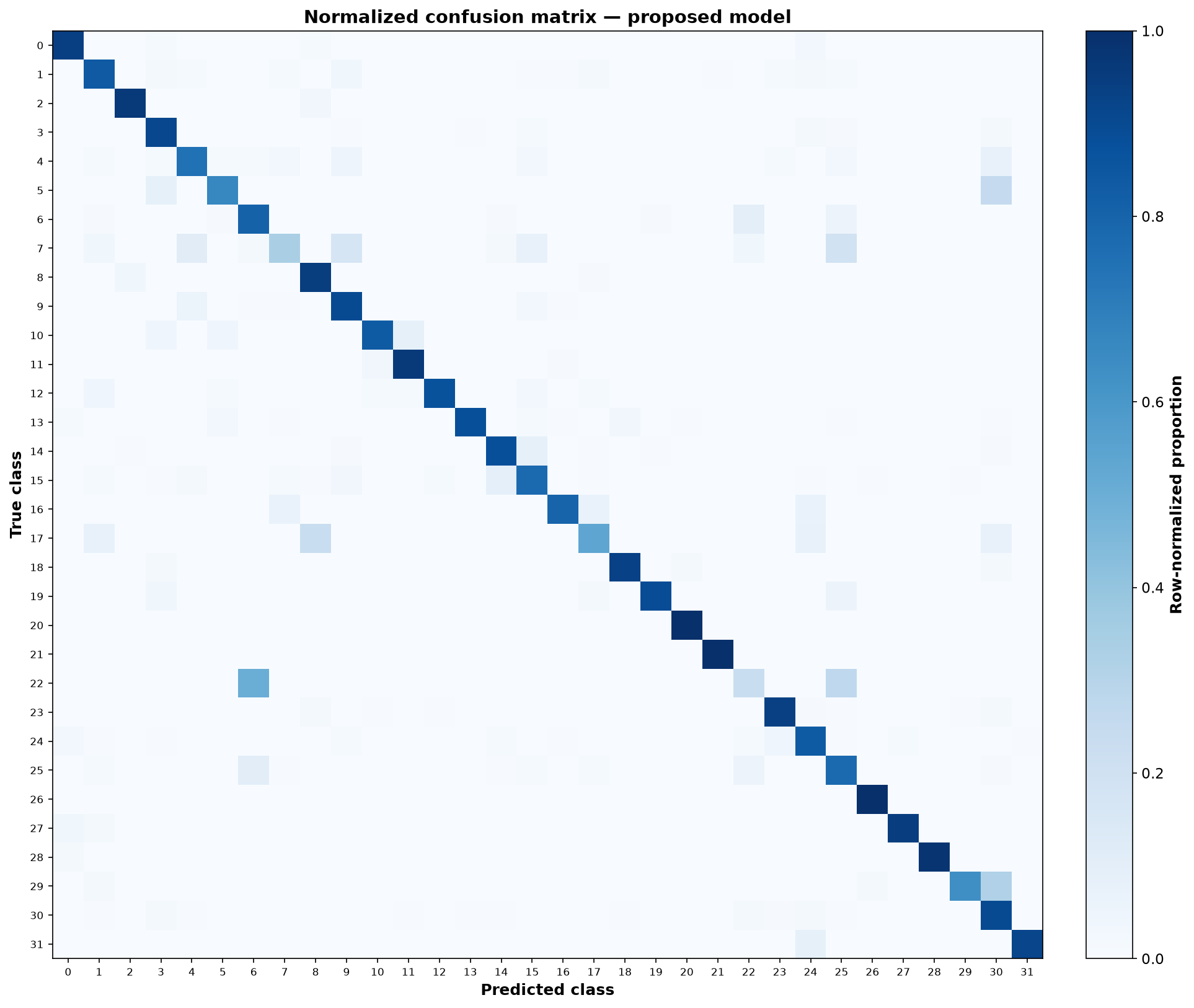}
\caption{Row-normalized 32-class confusion matrix.}
\end{subfigure}
\caption{Class-level evaluation of TAP-Path on the locked internal test set. The balanced panel layout combines class-wise F1 and normalized confusion behavior to expose performance heterogeneity that is not visible from aggregate accuracy alone.}
\label{fig:classbehavior}
\end{figure*}

The validation trade-off is shown in Fig.~\ref{fig:rareloss}, and class-level behavior is summarized in Fig.~\ref{fig:classbehavior}. When the selected logit-adjusted head was repeated across three seeds on the locked compressed encoder, test rare-class BA increased to $68.64\pm2.16$\%, balanced accuracy to $82.29\pm0.38$\%, while accuracy decreased to $87.13\pm0.68$\%. The comparison demonstrates a controllable operating trade-off within the same compressed encoder: rare-aware optimization increases minority-class sensitivity while shifting the balance among aggregate accuracy, balanced accuracy, and macro-F1. Figure~\ref{fig:classbehavior} further shows that the overall score is supported by strong discrimination in several well-represented classes: prostate adenocarcinoma reaches F1=0.996, testicular germ-cell tumors F1=0.986, thyroid carcinoma F1=0.986, kidney renal clear-cell carcinoma F1=0.963, and lower-grade glioma F1=0.961. The most difficult classes include rectum adenocarcinoma (F1=0.174), mesothelioma (0.378), esophageal carcinoma (0.439), and cholangiocarcinoma (0.552). The normalized confusion matrix confirms that errors are concentrated in a limited subset of class pairs rather than being uniformly distributed across the 32-class taxonomy. This heterogeneity motivates reporting macro-F1 and rare-class BA alongside top-1 accuracy. The primary paper result therefore remains the single-head mild-CE configuration in Table~\ref{tab:main}, with the rare-aware result used to explain the robustness/imbalance behavior of the architecture.

\begin{table*}[t]
\centering
\caption{Unified ablation study. The table intentionally combines structural, token, and rare-aware objective experiments rather than creating a separate headline table for the rare-aware operating point. Screening values are validation-only and should not be compared directly with final 12-patch test results.}
\label{tab:ablation}
\small
\begin{tabular}{llrrrrr}
\toprule
Component & Variant & Acc. (\%) & BA (\%) & Macro-F1 (\%) & Rare BA (\%) & FLOPs (G)\\
\midrule
\multirow{4}{*}{Structure} & Depth32 & 72.96 & 73.08 & 71.67 & -- & 340.13\\
& Depth24 & 68.28 & 68.38 & 67.30 & -- & 255.19\\
& TaskSparse22 & 69.18 & 69.30 & 68.01 & -- & 233.96\\
& \textbf{TaskSparse24} & \textbf{69.79} & \textbf{69.84} & \textbf{68.54} & -- & 255.19\\
\midrule
\multirow{3}{*}{Token ratio} & 100\% & 69.79 & 69.84 & 68.54 & -- & 255.19\\
& 85\% & 70.39 & 70.42 & 69.10 & -- & 237.71\\
& \textbf{70\%} & \textbf{71.60} & \textbf{71.64} & \textbf{70.32} & -- & \textbf{220.40}\\
\midrule
\multirow{5}{*}{Loss (validation)} & Mild CE & 89.99 & 82.82 & 84.47 & 67.39 & 220.40\\
& Balanced Softmax & 89.19 & 83.17 & 81.51 & \textbf{70.55} & 220.40\\
& Logit adjustment $\tau=.5$ & 89.94 & 83.04 & 84.24 & 67.74 & 220.40\\
& \textbf{Logit adjustment $\tau=1$} & \textbf{90.25} & \textbf{83.84} & 83.22 & 69.68 & 220.40\\
& CB focal $\gamma=1.5$ & 87.35 & 81.27 & 81.76 & 68.33 & 220.40\\
\midrule
Rare-aware 3-seed test & selected $\tau=1$ head & 87.13$\pm$.68 & 82.29$\pm$.38 & 80.74$\pm$.65 & 68.64$\pm$2.16 & 220.40\\
\bottomrule
\end{tabular}
\end{table*}

\subsection{External validation}
\label{sec:external}

External validation is reported as two distinct disease cohorts. CPTAC-CCRCC contains 209 independent tumor cases mapped to the internal kidney renal clear-cell carcinoma class, and CPTAC-UCEC contains 224 independent tumor cases mapped to uterine corpus endometrial carcinoma. No CPTAC image or label is used for architecture selection, head optimization, calibration, or threshold tuning.

Across the three primary single-head seeds, TAP-Path achieves 87.40$\pm$0.28\% accuracy on CCRCC and 94.79$\pm$1.44\% on UCEC. Pooled over 433 cases, accuracy is 91.22$\pm$0.83\% and present-class balanced accuracy is 91.10$\pm$0.81\%. Because each cohort corresponds to one mapped target class, cohort-specific recalls/accuracies can be recovered exactly from each frozen seed using
\begin{align}
\mathrm{BA}_{\mathrm{ext}} &= \frac{a_C+a_U}{2},\\
\mathrm{Acc}_{\mathrm{ext}} &=
\frac{209a_C+224a_U}{433}.
\end{align}

\begin{table}[t]
\centering
\caption{Frozen TAP-Path validation on the two external CPTAC cohorts. Values are mean$\pm$SD across the three primary single-head seeds.}
\label{tab:externalcohort}
\small
\begin{tabular}{lrr}
\toprule
Cohort & Cases & Accuracy (\%)\\
\midrule
CPTAC-CCRCC & 209 & \textbf{87.40$\pm$0.28}\\
CPTAC-UCEC & 224 & \textbf{94.79$\pm$1.44}\\
\midrule
Pooled CPTAC & 433 & \textbf{91.22$\pm$0.83}\\
Present-class BA & 433 & \textbf{91.10$\pm$0.81}\\
\bottomrule
\end{tabular}
\end{table}

\begin{figure}[t]
\centering
\includegraphics[width=\columnwidth]{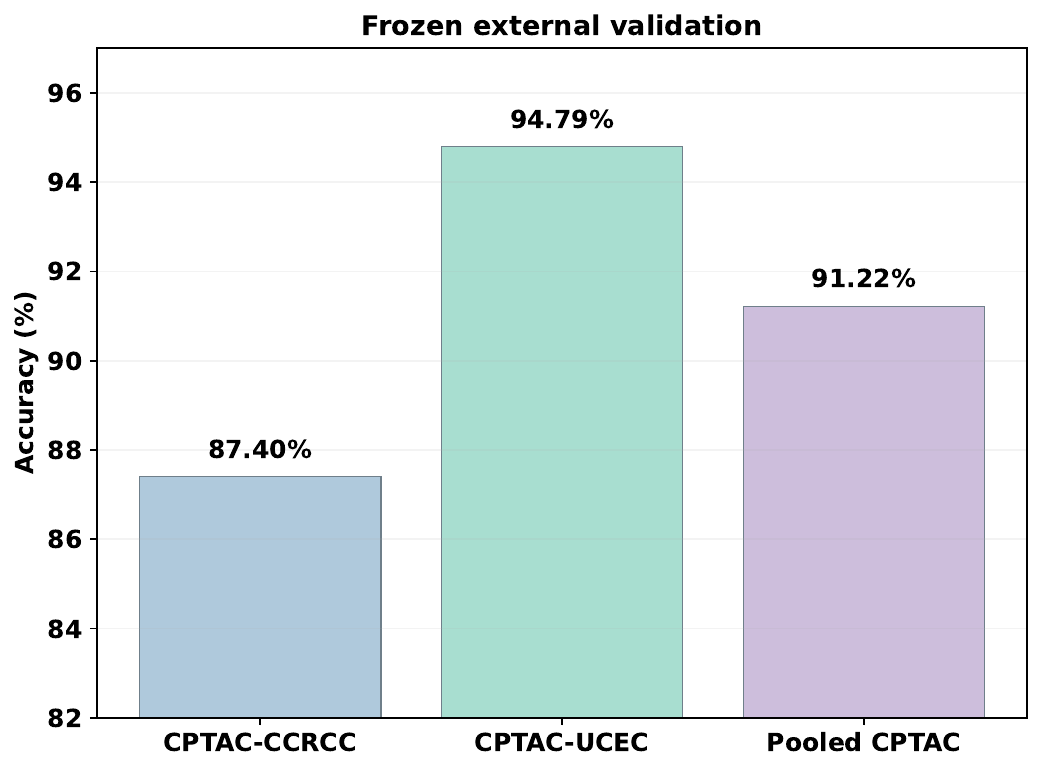}
\caption{Frozen external validation of TAP-Path on CPTAC-CCRCC and CPTAC-UCEC.}
\label{fig:externalrobust}
\end{figure}

Table~\ref{tab:externalcohort} reports the cohort-specific values and Fig.~\ref{fig:externalrobust} contrasts internal and external behavior. The cohort difference is meaningful: UCEC transfers more strongly than CCRCC under the locked model, demonstrating why external performance should not be collapsed into a single unqualified robustness claim. The result supports frozen transfer across these two morphologies, not universal 32-class external generalization.

Frozen external evaluation was performed on 433 CPTAC images after the internal architecture and task protocol were locked. Across three single-head seeds, \model achieved $91.22\pm0.83$\% accuracy and $91.10\pm0.81$\% balanced accuracy. External ECE was $0.0323\pm0.0086$ and failure-detection AUROC was $0.8974\pm0.0089$. These values indicate that the compressed representation remained useful outside the internal data source.

The external task is simpler in class cardinality than the internal 32-class benchmark, so its higher raw accuracy should not be interpreted as evidence that CPTAC is ``harder'' or that generalization improves with domain shift. Instead, the correct interpretation is that the locked model transferred successfully to the two externally represented classes. Among the retained baselines, Hibou-B obtained a slightly higher external accuracy (91.53\%), while \model exceeded \vir and UNI2-h. The external experiment therefore provides evidence of transfer robustness for the two represented classes.

\subsection{Representation analysis}

\begin{figure}[t]
\centering
\includegraphics[width=\columnwidth]{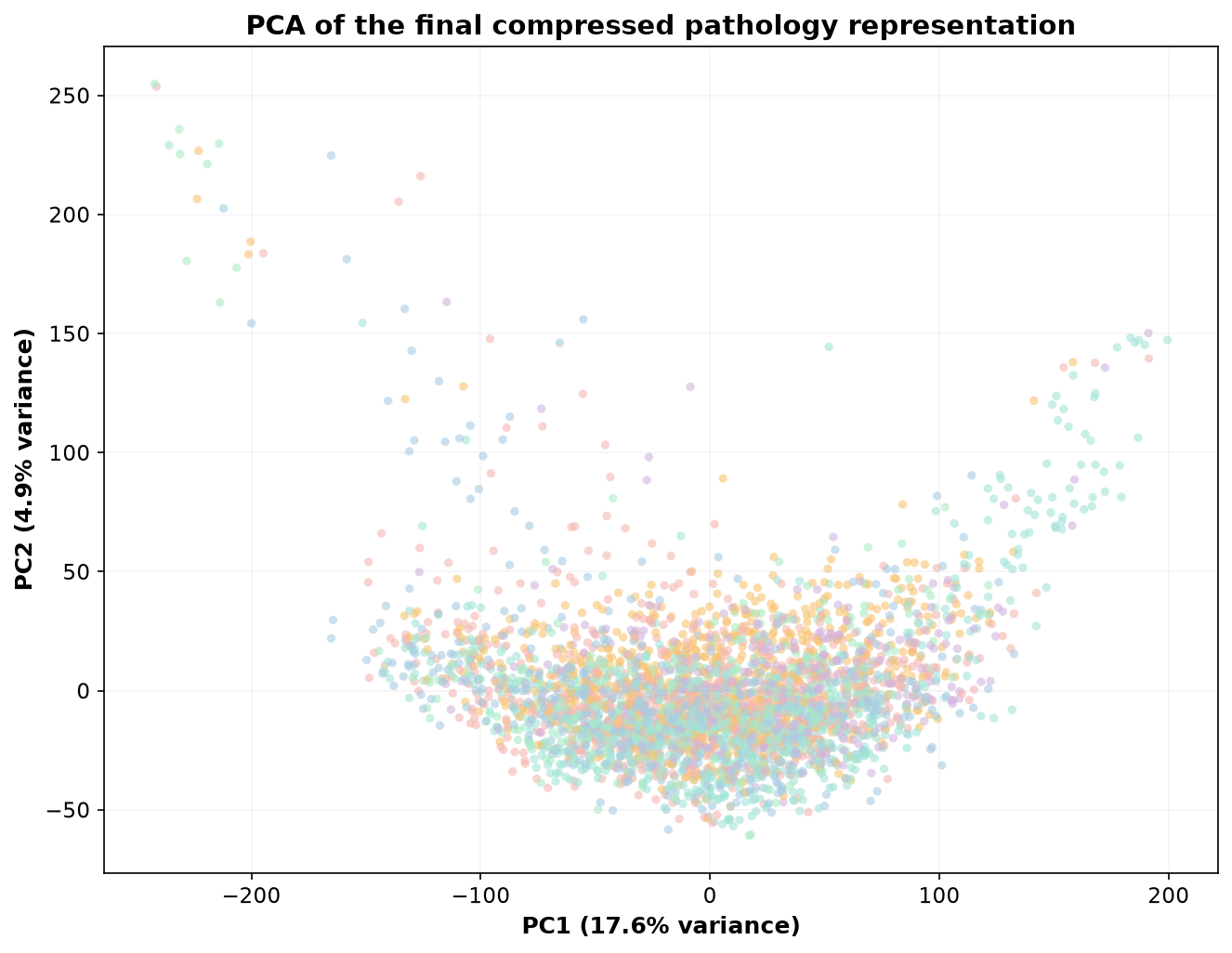}
\caption{PCA of held-out TAP-Path image-level representations.}
\label{fig:pca}
\end{figure}

As shown in Fig.~\ref{fig:pca}, PCA is used as a descriptive visualization of the locked compressed representation, not as a training signal. For standardized image-level feature matrix $Z$, the $k$th component is
\begin{equation}
w_k=\arg\max_{\|w\|_2=1,\;w\perp w_{<k}}\operatorname{Var}(Zw).
\end{equation}
The projection in Fig.~\ref{fig:pca} shows that broad class-dependent structure remains visible after depth and token reduction, although substantial overlap is expected for a 32-class pan-cancer task.

\subsection{Deployment efficiency}

The final compressed encoder has 473.70M parameters and the single task head has 5.70M, giving 479.40M deployed parameters. Analytical encoder compute is 220.40G FLOPs. An independent profiler supported approximately 213.98G FLOPs for the measured execution path. On the NVIDIA GeForce RTX 5060 Ti development system, mean inference latency was 31.85 ms with 1.21 ms standard deviation, corresponding to 31.40 images/s. These numbers are hardware-specific and should not be generalized to clinical scanners or server deployments, but they confirm that the structural and token changes translate into an executable compressed model rather than only theoretical masking.

\subsection{Bootstrap uncertainty}

\begin{figure}[t]
\centering
\includegraphics[width=\columnwidth]{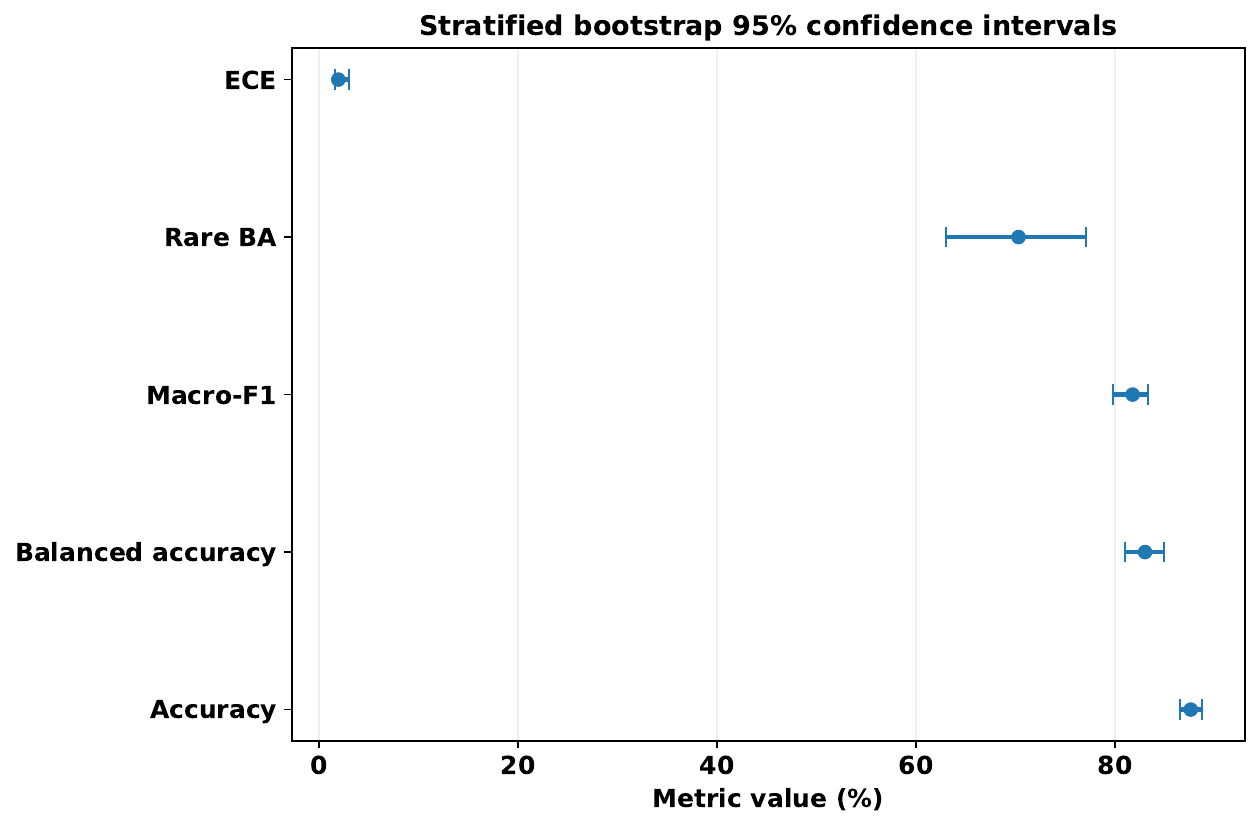}
\caption{Stratified 95\% bootstrap confidence intervals for the locked TAP-Path test evaluation.}
\label{fig:bootstrap}
\end{figure}

Figure~\ref{fig:bootstrap} summarizes the resulting uncertainty. The 2,000-resample stratified bootstrap gives approximate 95\% intervals of 0.870--0.889 for accuracy, 0.793--0.831 for balanced accuracy, 0.805--0.839 for macro-F1, 0.568--0.698 for rare-class balanced accuracy, and 0.018--0.032 for ECE. The wider rare-class interval reflects the smaller effective support of low-frequency classes.

\section{Discussion}

\subsection{Whole-slide efficiency}

The reported FLOP reduction is measured per encoder invocation, but digital pathology multiplies this saving across patch bags. Although the present experiments evaluate fixed image-level patch bags rather than end-to-end WSI inference, the per-patch analytical FLOP reduction can be extrapolated to illustrate potential whole-slide computational savings. If a WSI contributes $N_{\mathrm{WSI}}$ encoded patches,
\begin{equation}
\Phi_{\mathrm{slide}}\approx N_{\mathrm{WSI}}\Phi_{\mathrm{patch}}.
\end{equation}
For 10,000 encoded patches, reducing analytical cost from 340.13G to 220.40G FLOPs corresponds to roughly $1.20\times10^{15}$ fewer floating-point operations before downstream aggregation. This is a theoretical compute estimate rather than a latency guarantee, yet it illustrates why a 35.20\% per-patch reduction is meaningful at WSI scale.

\subsection{Structural compression rationale}

The structural compression results demonstrate that \model establishes a favorable accuracy--efficiency operating point derived directly from a high-capacity pathology foundation model. Relative to the full \vir backbone, \model removes approximately one quarter of the encoder parameters and more than one third of the analytical encoder FLOPs while maintaining strong predictive performance. Importantly, this reduction is achieved through task-adaptive physical reconstruction of the encoder rather than through masking or freezing redundant components. The resulting architecture therefore preserves the representational advantages of large-scale pathology pretraining while reducing the computational burden associated with downstream inference. The concurrent improvements in test accuracy and macro-F1 further indicate that the removed structural components are not essential for the target classification task and that task-adaptive compression can produce a more efficient downstream representation without sacrificing predictive utility.

This property distinguishes \model from parameter-efficient fine-tuning approaches. Adapter- and prompt-based methods can substantially reduce the number of parameters updated during task adaptation, and approaches such as PAMT demonstrate the effectiveness of this strategy in computational pathology \cite{lin2026pamt}. However, these methods primarily reduce optimization cost while retaining most of the pretrained backbone for inference. In contrast, \model directly reconstructs a smaller executable encoder by removing task-redundant transformer blocks, thereby reducing both deployed parameter count and analytical inference cost. This distinction is particularly relevant for pathology applications in which large numbers of image regions may need to be processed repeatedly and inference efficiency becomes an important deployment consideration.

Knowledge-distillation approaches provide another complementary route toward model compression. Methods such as SUDA transfer knowledge from a larger teacher into a compact student architecture \cite{zhong2026suda}, enabling substantial reductions in model size through an additional teacher--student optimization stage. \model addresses a different compression objective: rather than replacing the pretrained foundation model with a separately trained student, it identifies and preserves the task-relevant structure within the pretrained model itself. Consequently, structural pruning, parameter-efficient adaptation, and knowledge distillation represent complementary strategies operating at different stages of the model lifecycle. Their combination provides a promising direction for future work, particularly through hybrid pruning--distillation frameworks that first remove task-redundant structure and subsequently transfer the retained knowledge into an even smaller deployment model.

\subsection{Non-contiguous depth selection}

The structural screen provides evidence that depth alone is not an adequate proxy for task relevance. At equal 24-block parameter count, TaskSparse24 outperformed contiguous Depth24 during validation screening. The result suggests that useful transformations are distributed non-uniformly through the pretrained hierarchy and that preserving selected late/middle blocks can recover information lost by naive truncation. This is consistent with the broad observation that pretrained transformer representations change qualitatively across depth.

We deliberately anchored early and final blocks. The early anchor protects low-level patch/token formation, while the late anchor preserves semantic consolidation close to the original pretrained output. The middle selection is then allowed to adapt. Although the novelty score in Eq.~\eqref{eq:novelty} is simple, its practical advantage is that it is model-internal, label-efficient for profiling, and compatible with physical block removal. A limitation is that normalized residual change is not a causal attribution score; a block can have a small residual magnitude yet still be important through subtle feature refinement. Future work could compare the current score with gradient- or Hessian-based block saliency.

\subsection{Token-retention behavior}

The token ablation is particularly useful because the selected 70\% ratio simultaneously improved screening validation accuracy and reduced compute for TaskSparse24. One plausible interpretation is that class-token similarity plus magnitude suppresses redundant or weakly aligned patches after sufficient hierarchical processing. Similar motivations underlie task-aware token pruning in generic vision transformers \cite{liu2024tokenpruning,bergner2025tokencropr}. However, this effect was not universal: TaskSparse22 degraded as tokens were removed. Token sparsity therefore interacts with representational depth and should be selected jointly rather than treated as an independent compression knob.

In pathology, this interaction deserves caution. Small discriminative regions can be clinically meaningful, so a high compression ratio could remove uncommon morphology. Our 70\% setting was chosen by a validation Pareto rule and was not tuned on the test set. Moreover, rare-class behavior was evaluated explicitly. These safeguards reduce but do not eliminate the risk of token-pruning bias.

\subsection{Reliability interpretation}

The reliability analysis shows that the compressed model preserves informative probabilistic behavior alongside its efficiency gains. The single-head \model achieved the best Brier score and failure-detection AUROC among the directly compared large-model configurations, while ECE and NLL remained close to UNI2-h. A failure-detection AUROC of approximately 0.90 indicates effective ranking of incorrect predictions toward lower confidence, supporting confidence-guided selective review. Clinical operating thresholds and safety characteristics require prospective, workflow-specific evaluation beyond these statistical reliability measures.

Rare-class analysis further characterizes how the optimization objective changes model behavior. The primary objective favors aggregate predictive performance, whereas the rare-aware head increases rare-class balanced accuracy with a corresponding shift in overall accuracy. Reporting the two operating points separately provides a direct view of the accuracy--minority-sensitivity trade-off and preserves the metric provenance of each configuration.

\subsection{External transfer}

Independent evaluation is increasingly considered essential for pathology foundation models \cite{campanella2025benchmark,neidlinger2025benchmark,komen2026robust}. The CPTAC results show that \model maintains high accuracy and balanced accuracy after freezing the internal configuration. The difference in class cardinality prevents direct comparison of internal 32-class macro-F1 with an external two-class macro-F1, which is why the manuscript emphasizes present-class balanced accuracy externally.

Across the independent CPTAC evaluation, \model maintained strong transfer performance after the internal configuration was frozen. Hibou-B achieved slightly higher raw CPTAC accuracy, consistent with prior evidence that foundation-model rankings vary across domains and tasks \cite{neidlinger2025benchmark,bareja2026benchmark}. For the studied 32-class task, the combined internal and external results position \model as an efficient high-capacity configuration with stable transfer behavior under the evaluated CPTAC cohorts.

\subsection{Relation to efficient pathology methods}

Recent pathology-efficiency studies provide important context for this work. SUDA achieves aggressive compression by transferring knowledge from a large teacher to a compact student \cite{zhong2026suda}, whereas PAMT reduces downstream computational requirements through representative patch selection and lightweight adaptation of frozen representations \cite{lin2026pamt}. Other studies have explored resource-efficient architectures and accuracy--complexity trade-offs for histopathology analysis \cite{islam2025involution,lee2025benchmark}. Despite these advances, efficiency is commonly achieved through student-model distillation, input-level reduction, lightweight adaptation, or architectures specifically designed for efficient inference. \model addresses a complementary setting: task-adaptive compression of an already pretrained pathology foundation model. Rather than pretraining a new foundation model, distilling a separate student, or retaining the complete parent architecture, \model restructures the pretrained parent for the downstream task by preserving task-relevant depth, reducing redundant token computation, and recovering complementary intermediate representations within a single compressed inference pathway.

Accordingly, \model is evaluated through a joint accuracy--efficiency--reliability perspective that integrates predictive performance with computational cost and probabilistic behavior. The proposed framework preserves competitive predictive performance relative to substantially larger pathology foundation models while reducing both model size and computational demand. This efficiency is evaluated together with three-seed reproducibility, probability calibration, failure-awareness analysis, and frozen external validation. Collectively, these evaluations examine whether a pretrained pathology foundation model can be structurally compressed for a specific downstream task while retaining predictive performance, computational efficiency, and informative confidence behavior.

\section{Limitations}

Several directions remain for extending the present evaluation. First, \model contains 479.40M deployed parameters, placing it between compact encoders such as Hibou-B/CONCH and larger \vir/UNI2-h systems. Further compression could target lower-resource deployment settings. Second, analytical FLOPs provide a reproducible architecture-level efficiency measure, while end-to-end deployment also depends on hardware, energy consumption, WSI throughput, slide loading, tissue detection, patch extraction, and storage I/O. The measured 31.85 ms image-level latency therefore characterizes the evaluated hardware configuration and motivates broader systems-level benchmarking.

Third, the current external evaluation covers CCRCC and UCEC, providing an independent transfer assessment for two classes represented in the internal taxonomy. Extending this protocol to additional cancer types, scanners, and institutions would enable broader pan-cancer robustness analysis. Fourth, the primary objective prioritizes aggregate performance, while the rare-aware analysis demonstrates that minority-class sensitivity can be increased within the same compressed encoder. This motivates future structural-selection criteria that incorporate class-frequency or minority-utility information directly during compression.

Fifth, the block-novelty score quantifies normalized representational change and serves as a computationally efficient task-selection criterion. Future work can compare this criterion with gradient-, Hessian-, or attribution-based structural importance measures and examine how retained-block patterns transfer across endpoints. Finally, the present reliability analysis is statistical and model-centered; prospective clinical evaluation, reader studies, workflow-specific decision thresholds, and formal safety assessment are natural next steps for establishing clinical utility.

\section{Conclusion}

We presented \model, a task-adaptive structural and token pruning framework for converting a large pretrained pathology foundation model into a more efficient task-specific encoder without knowledge distillation. Starting from \vir, the method profiles task-relevant block novelty, physically retains 24 of 32 transformer blocks, adaptively preserves 70\% of patch tokens after the pruning point, and recovers multi-depth information through a gated single task head. The resulting encoder reduces parameters by 24.96\% and analytical FLOPs by 35.20\%. Across three independent single-head runs, \model achieves 87.98$\pm$0.067\% internal test accuracy and 82.38$\pm$0.48\% macro-F1, matching or slightly exceeding much larger \vir and UNI2-h baselines on the studied task. The compressed model also preserves useful probabilistic behavior, including a Brier score of 0.1800 and failure-detection AUROC of 0.9047, while frozen CPTAC evaluation reaches 91.22$\pm$0.83\% accuracy.

The results support a practical conclusion: scaling pathology foundation models and compressing them for deployment are not mutually exclusive. A large pretrained hierarchy can contain downstream-task redundancy that is removable through validation-constrained structural and token sparsification. At the same time, compression should be evaluated beyond top-line accuracy. Explicit reporting of calibration, error awareness, rare-class trade-offs, and independent external behavior is important for assessing whether the resulting efficiency gain remains suitable for trustworthy medical-AI research.

\section*{Data Availability}

The histopathology data analyzed in this study were derived from publicly available resources from The Cancer Genome Atlas (TCGA) \cite{tcga2013pancancer} and the Clinical Proteomic Tumor Analysis Consortium (CPTAC) \cite{clark2019ccrcc,dou2020ucec}. TCGA data are accessible through the NCI Genomic Data Commons, while the corresponding CPTAC resources are available through NCI-supported data repositories. No external evaluation data were used for architecture selection or model optimization.






\appendix
\section{Reproducibility details}

\subsection{Primary training configuration}
\begin{table}[h]
\centering
\caption{Primary training configuration.}
\small
\begin{tabular}{ll}
\toprule
Setting & Value\\
\midrule
Parent foundation model & Virchow2\\
Retained blocks & 24/32\\
Patch-token retention & 70\%\\
Multi-depth taps & 4\\
Tap projection & 256\\
Dropout & 0.18\\
Optimizer & AdamW\\
Learning rate & $7\times10^{-4}$\\
Weight decay & $2\times10^{-4}$\\
Batch size & 384\\
Maximum epochs & 130\\
Early stopping & 18 epochs\\
Gradient clip & 5\\
Label smoothing & 0.01\\
Seeds & 42, 123, 2026\\
Calibration & validation-only temperature scaling\\
\bottomrule
\end{tabular}
\end{table}

\subsection{Experimental provenance}
The manuscript separates architecture screening, primary task evaluation, common-probe diagnostics, rare-aware operating-point analysis, and external validation. Metrics are not silently transferred between these protocols. This distinction is essential because the common-probe analysis answers a representation-quality question, whereas the original TAP-Path task head defines the proposed deployed system.

\printbibliography

\end{document}